\documentclass{article}

\usepackage{tcolorbox}

 \usepackage[eandd, final]{neurips_2026}

\usepackage[utf8]{inputenc} 
\usepackage[T1]{fontenc}    
\usepackage{hyperref}       
\usepackage{url}            
\usepackage{booktabs}       
\usepackage{amsfonts}       
\usepackage{nicefrac}       
\usepackage{microtype}      
\usepackage{xcolor}         
\usepackage{pifont}
\usepackage{multirow}
\usepackage{graphicx}
\usepackage{subcaption}
\usepackage{wrapfig}
\usepackage{amsmath}
\usepackage{colortbl}
\usepackage[table]{xcolor}
\usepackage{makecell}
\usepackage{arydshln}
\usepackage{enumitem}

\definecolor{bestgreen}{RGB}{223,237,223}
\definecolor{secondred}{RGB}{243,224,228}
\definecolor{groupgray}{RGB}{120,120,120}

\title{Benchmarking and Learning Object-Centric Procedural Understanding in Instructional Videos}

\author{Wenliang Guo \\
Michigan State University, USA\\
\texttt{guowenli@msu.edu} \\
\And 
Yu Kong \\
Michigan State University, USA\\
\texttt{yukong@msu.edu} \\
}

\begin{document}

\maketitle

\begin{abstract}
Procedural activities are fundamentally driven by object state transitions, yet existing instructional video benchmarks remain action-centric and cannot evaluate whether models reason about how objects evolve toward task completion. In this work, we introduce \textsc{ProcObject-10K}, the first benchmark that jointly evaluates object-centric reasoning and temporal evidence grounding in instructional videos, across both egocentric and exocentric views. It comprises 10,522 open-ended VideoQA pairs grounded in 1,799 videos, spanning 137 tasks across 9 domains and five reasoning types covering preconditions, state evolution, counterfactuals, mistakes, and readiness. Benchmarking 13 leading MLLMs reveals a substantial answering-grounding gap: models produce plausible answers while failing to localize the supporting evidence (mIoU $<$ 45\%), exposing their reliance on linguistic priors rather than recognizing and reasoning about the object dynamics. As a step toward closing this gap, we further provide an object-centric supervised fine-tuning baseline with pseudo object-level supervision and spatial-temporal constraints. Models fine-tuned on \textsc{ProcObject-10K} not only improve on the benchmark itself, but also transfer effectively to other grounded VideoQA and embodied planning tasks. The dataset, annotations, and evaluation toolkit have been publicly released on our project page to support future research on object-centric procedural understanding: \url{https://wenliangguo.github.io/ProcObject-10K-Page/}.
\end{abstract}    
\section{Introduction}
\label{sec:intro}

Human daily activities are inherently procedural, spanning scenarios such as cooking \cite{damen2020epic,peddi2024captaincook4d}, assembly \cite{sener2022assembly101}, and medical procedures \cite{ozsoy2025egoexor}. These activities are governed by an underlying \emph{procedural structure}, where actions operate on objects to induce state transitions, forming causal dependencies across steps toward task completion. For example, cracking an egg enables mixing, and mixing produces a liquid state required for subsequent cooking. Understanding such structure requires not only recognizing actions \cite{Bao_2021_ICCV}, but reasoning about how object states evolve over time \cite{souvcek2022look} and how these changes constrain future actions. This capability is fundamental for future intelligent systems that aim to assist humans on procedural tasks by watching and learning from instructional videos.

Existing work on procedural understanding predominantly adopts an action-centric perspective. Representative approaches construct symbolic graphs over action sequences to model dependencies \cite{jang2023multimodal, ashutosh2023video, soran2015generating, sohn2020meta, Huang_2025_CVPR, lee2025error}, or rely on masked modeling to recover missing action segments and implicitly learn structure \cite{narasimhan2023learning, lin2024vedit, zhong2023learning}. Despite methodological differences, these approaches share a common assumption that procedural structure is determined by action transitions alone.

However, this assumption overlooks a key property of procedural activities whose goal is achieved through progressive transformation of object states under human interaction. Procedural causality is therefore not fully captured by action sequences, but is also reflected in how object states evolve over time. Direct evidence is that the same action can produce different outcomes depending on execution conditions \cite{guo2026procedural}, which in turn alters the set of valid subsequent actions \cite{niu2024schema}. As a result, action-centric representations of procedural structure lack sufficient object-state awareness, limiting their ability to support consistent reasoning about procedural progress and causal dependencies.

\begin{table*}[t]
\renewcommand{\arraystretch}{1.1}
\centering
\caption{Dataset comparison. \textsc{ProcObject-10K} provides a comprehensive evaluation of open-ended question answering, temporal evidence grounding, object-centric reasoning, and mistake understanding across both egocentric and exocentric instructional videos.}
\footnotesize
\setlength{\tabcolsep}{4.5pt}
\resizebox{\linewidth}{!}{%
\begin{tabular}{lcccccc}
\toprule
\textbf{Datasets} & \textbf{View} & \textbf{Object-centric} & \textbf{QA Type} & \textbf{Evidence} & \textbf{Mistake} & \textbf{Video Domain} \\
\midrule
COIN \cite{tang2019coin} & \underline{Exo} & \ding{55} & - & \ding{55} & \ding{55} & Instructional \\
ChangeIt \cite{souvcek2022look} & \underline{Exo} & \ding{51} & - & \ding{55} & \ding{55} & Instructional \\
EgoSchema \cite{mangalam2023egoschema} & Ego & \ding{55} & Multi-choice & \ding{55} & \ding{55} & Human Activity \\
EgoPER \cite{lee2024error} & Ego & \ding{55} & - & \ding{55} & \ding{51} & Instructional \\
CaptainCook4D \cite{peddi2024captaincook4d} & Ego & \ding{55} & - & \ding{55} & \ding{51} & Instructional \\
ProMQA \cite{hasegawa2024promqa} & Ego & \ding{55} & Open-ended & \ding{55} & \ding{51} & Instructional\\
REXTIME \cite{chen2024rextime} & \underline{Exo} & \ding{55} & Multi-choice & \ding{51} & \ding{55} & Generic \\
VideoInfer \cite{wang2025object} & \underline{Exo} & \ding{51} & Open-ended & \ding{55} & \ding{55} & Generic \\
\textsc{MultiHop-EgoQA} \cite{chen2025grounded} & Ego & \ding{55} & Open-ended & \ding{51} & \ding{55} & Human Activity \\
TrackVerse \cite{wei2025trackverse} & \underline{Exo} & \ding{51} & - & \ding{55} & \ding{55} & Generic \\
EPIC-KITCHENS-100-MQA \cite{qi2026llavaction} & Ego & \ding{55} & Multi-choice & \ding{55} & \ding{55} & Instructional \\
\textbf{\textsc{ProcObject-10K}} & Ego+\underline{Exo} & \ding{51} & Open-ended & \ding{51} & \ding{51} & Instructional \\
\bottomrule
\end{tabular}%
}
\label{tab:dataset}
\end{table*}

\begin{figure}[t]
    \centering
    \includegraphics[width=1\linewidth]{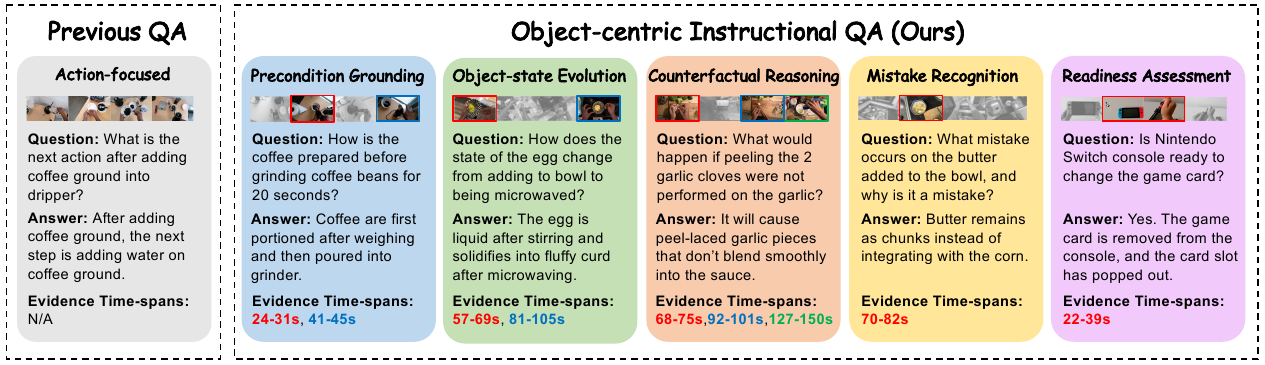}
    \caption{QA types in \textsc{ProcObject-10K} dataset, which centers on object-level understanding to capture temporal and spatial dynamics, enabling benchmarking from diverse perspectives.}
    \label{fig:qa_type}
\end{figure}

In this work, we study the procedural understanding of instructional videos from an object-centric perspective, where object state dynamics serve as an observable signal for modeling procedural structure. We formulate a procedure as a sequence of temporally grounded object state transitions with localized evidence and precondition constraints, enabling explicit evaluation of whether models can identify relevant objects, localize state changes, and reason about their causal roles. However, previous benchmarks fail to support such evaluation, because existing instructional video datasets \cite{tang2019coin, lee2024error, peddi2024captaincook4d, hasegawa2024promqa, qi2026llavaction} focus on action sequences without modeling object state changes, while object-centric datasets \cite{souvcek2022look, wang2025object, wei2025trackverse} lack goal-driven procedural structure or object-level causal dependencies. Consequently, current benchmarks fail to evaluate whether models reason over object state dynamics, instead allowing strong performance through reliance on action patterns.

To bridge this gap, we introduce \textsc{ProcObject-10K}, an instructional video benchmark for object-centric procedural understanding. It is formulated as a grounded VideoQA task \cite{di2023groundvqa, chen2025grounded, xiao2024can, gupta2025toga} where each sample requires reasoning over temporally localized evidence of object state changes. The dataset contains 1{,}799 videos and 10,522 question-answer pairs with aligned evidence spans, covering 137 diverse tasks across 9 domains. It is constructed via a semi-automated pipeline with VLM generated annotations followed by model-based and manual verification and refinement for temporal and causal consistency. To systematically evaluate procedural reasoning, we define five types of questions shown in Figure \ref{fig:qa_type}: Precondition Grounding, Object State Evolution, Counterfactual Reasoning, Mistake Recognition, and Readiness Assessment, covering complementary aspects of procedural structures. Our benchmarking results expose a critical Answering-Grounding Gap: while leading MLLMs achieve plausible performance in language answering, they consistently fail to pinpoint the underlying temporal evidence, with grounding mIoU generally remaining below 45\%. This discrepancy reveals that existing models often rely on linguistic priors rather than achieving an object-centric understanding of procedural evolution.

To further validate the diagnostic value of the benchmark, we introduce an object-centric supervised fine-tuning (SFT) framework that explicitly encourages MLLMs to attend to the dynamic evolution of action-relevant objects. To this end, we construct object-level pseudo labels using an object grounding model \cite{liu2023grounding} and a vision-language model \cite{qwen3.5}, and apply spatial and temporal attention constraints as supervision signals to guide models toward relevant objects and key frames. This training strategy improves both grounding accuracy and reasoning performance, suggesting that explicit supervision on object state dynamics addresses the failure modes exposed by the benchmark.

Experimental results show that the learned object-centric representation not only improves performance on \textsc{ProcObject-10K}, but also exhibits strong generalization. The model fine-tuned on \textsc{ProcObject-10K} transfers effectively to other procedural VideoQA benchmarks \cite{chen2025grounded} under zero-shot settings, and achieves improved performance on embodied instruction following tasks \cite{kim2025multimodal, ALFRED20} when used as zero-shot planners. These results indicate that reasoning over object state dynamics yields a transferable capability that extends beyond the object-centric benchmark setting, highlighting the broader value of \textsc{ProcObject-10K} for procedural understanding.

In summary, our contributions are as follows:
\begin{itemize}[leftmargin=0.5cm]
\vspace{-2mm}
\item We introduce \textsc{ProcObject-10K}, a benchmark for object-centric procedural understanding, featuring 10{,}522 VideoQA pairs grounded in 1{,}799 videos across 137 tasks and 9 domains.
\item We propose an SFT framework that incorporates object-level supervision and spatial-temporal constraints to implicitly model procedural structure and learn object-centric representation.
\item We empirically show that object-centric understanding not only improves procedural reasoning and grounding, but also generalizes to other  VideoQA and even embodied planning tasks.
\vspace{-2mm}
\end{itemize}
\section{Related Work}
\label{sec:related_work}

\noindent\textbf{Procedural Understanding.}
Instructional videos depicting goal-driven activities such as cooking \cite{damen2020epic, peddi2024captaincook4d, lee2024error} and assembly \cite{sener2022assembly101, hasegawa2025promqa} have been widely studied for procedural understanding. Prior work has primarily focused on action-centric tasks, including action recognition \cite{Bao_2021_ICCV}, action segmentation \cite{zhang2022actionformer}, and action anticipation \cite{gong2022future}. Beyond procedural understanding, recent studies in general video understanding have explored object-centric representations through finer-grained modeling of object dynamics and state transitions \cite{yan2024visa, wei2025trackverse, wang2025object}. Since procedural activities inherently involve dense human-object interactions, object-centric modeling has also shown strong potential for instructional video understanding, benefiting downstream tasks such as object state prediction \cite{zameni2025moscato}, procedural planning \cite{niu2024schema}, and mistake detection \cite{guo2026procedural}. These works highlight the importance of modeling object dynamics for procedural reasoning. However, existing approaches are typically designed for specific downstream tasks \cite{guo2026procedural, niu2024schema} or only capture short-term local object state changes \cite{souvcek2022look}, lacking a general capability for understanding procedural structure from an object-centric perspective. These limitations motivate the development of foundation models capable of object-centric procedural reasoning across more applications in this work.

\noindent\textbf{Video Question Answering.}
VideoQA has emerged as an effective task for evaluating spatial-temporal reasoning over videos through natural language interaction \cite{jang2017tgif,yu2019activitynet}. More recent grounded VideoQA benchmarks further require models to localize supporting temporal evidence for the answers, enabling evaluation beyond language generation toward evidence-aware reasoning \cite{di2023groundvqa,chen2025grounded,xiao2024can,gupta2025toga}. Existing benchmarks have advanced long-form reasoning \cite{mangalam2023egoschema} and temporal causal understanding \cite{chen2024rextime}, while instructional VideoQA datasets such as ProMQA \cite{hasegawa2024promqa,hasegawa2025promqa} and \textsc{MultiHop-EgoQA} \cite{chen2025grounded} study procedural reasoning over human-object interactions. However, these benchmarks primarily focus on action sequences or event-level reasoning, without explicitly modeling the causal dependencies induced by object state transitions across procedural steps. In contrast, we introduce a grounded VideoQA benchmark in this work for object-centric procedural understanding that explicitly evaluates object state evolution, temporal causal structure, and evidence localization in instructional videos.
\section{\textsc{ProcObject-10K} Benchmark}
\label{sec:dataset}

\begin{figure}[t]
    \centering
    \begin{subfigure}[t]{0.59\linewidth}
        \centering
        \includegraphics[width=\linewidth]{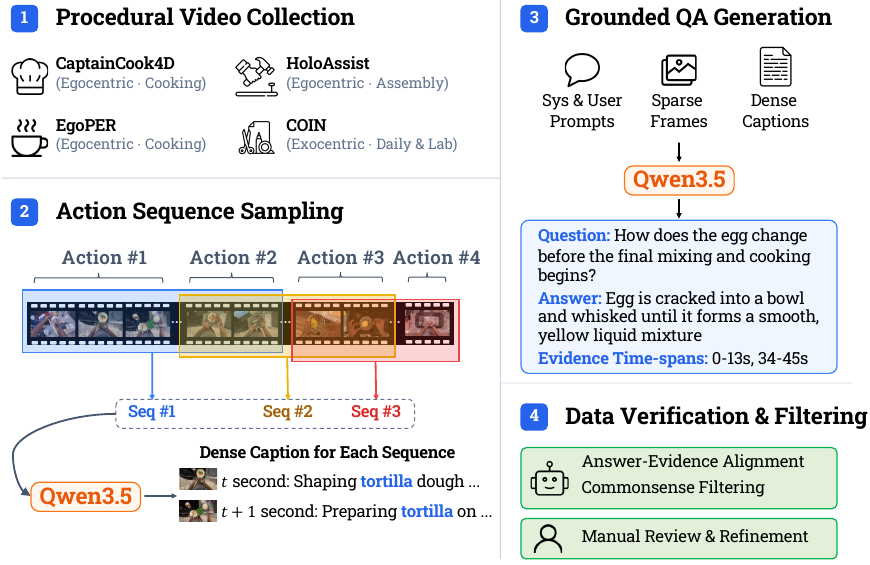}
        \caption{Data generation pipeline.}
        \label{fig:data_pipelne}
    \end{subfigure}
    \hfill
    \begin{subfigure}[t]{0.39\linewidth}
        \centering
        \includegraphics[width=\linewidth]{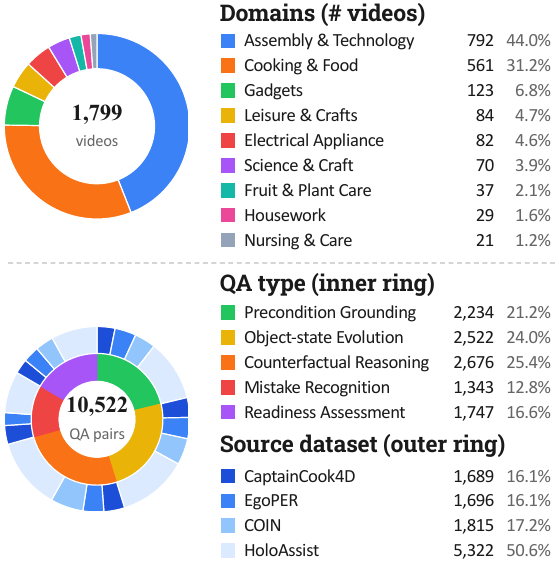}
        \caption{Dataset statistics. }
        \label{fig:data_statistic}
    \end{subfigure}
    \caption{Overview of \textsc{ProcObject-10K}: (a) data generation pipeline, and (b) distribution of task domains (upper chart) and distribution of video sources and QA types (lower chart). }
    \vspace{-6pt}
    \label{fig:data}
\end{figure}

Existing benchmarks fail to jointly capture object state dynamics and the underlying procedural structure. To bridge this gap, we introduce \textsc{ProcObject-10K}, a benchmark dedicated to object-centric procedural reasoning. It features open-ended, temporally grounded QA pairs that explicitly require models to localize relevant objects and infer causal state transitions across multiple execution steps. In this section, we first detail the data generation pipeline in Section \ref{sec:data_generate}, and then analyze the dataset statistics in Section \ref{sec:data_statistic}.

\subsection{Data Generation}
\label{sec:data_generate}

We construct the \textsc{ProcObject-10K} benchmark dataset using a four-stage semi-automated pipeline, as illustrated in Figure \ref{fig:data_pipelne}. To scale the benchmark dataset, we first collect instructional videos based on existing datasets. Then we process the raw videos and sample action sequences to maintain consistent object appearance and change while preserving procedural structure. We employ vision-language models \cite{qwen3.5} to generate QA pairs via multimodal prompting. Finally, we perform model-based and manual post verification and filtering to ensure data quality.

\noindent\textbf{Stage 1: Procedural Video Collection.}
We curate videos from datasets characterized by complex hand-object interactions, including the egocentric cooking datasets CaptainCook4D \cite{peddi2024captaincook4d} and EgoPER \cite{lee2024error}, as well as the egocentric assembly dataset HoloAssist \cite{HoloAssist2023}. In particular, these three datasets contain instances of erroneous actions and procedural failures, such as object mishandling or omission of critical steps, which provide a natural testbed to evaluate mistake detection \cite{lee2024error, lee2025error, guo2026procedural, patsch2025mistsense}. This capability is essential for assessing whether models capture the underlying procedural structure beyond surface level action recognition. To further broaden procedural coverage, we incorporate COIN \cite{tang2019coin}, a large scale Internet video dataset spanning diverse domains including daily activities, industrial operations, and scientific experiments. We apply a set of video filtering criteria to ensure data quality, including the removal of corrupted videos, elimination of highly similar procedures, and exclusion of overly long videos ($\geq$ 30 minutes). The resulting collection comprises 1,943 videos (before QA filtering in Stage 4) and serves as a basis for subsequent data generation.

\noindent\textbf{Stage 2: Action Sequence Sampling.}
Untrimmed procedural videos involve multiple interacting objects, making them unsuitable for directly constructing questions about specific object-state dynamics. To ensure that each video captures meaningful and causally coherent state transitions, we sample temporally contiguous action sequences.  Concretely, we segment each video into consecutive action clips and apply a temporal sliding window over $N$ segments to extract sub-action sequences, e.g., “Place tortilla on a cutting board $\rightarrow$ Pour egg mixture on tortilla $\rightarrow$ Roll the tortilla.” To balance object consistency while preserving rich state transitions, we set $N \in \{2,3,4,5\}$, enabling windows of varying lengths to yield diverse procedural sequences. To obtain object information, we prompt Qwen3.5 \cite{qwen3.5} to generate dense captions conditioned on action annotations, focusing on key objects such as the tortilla. These captions capture fine-grained object states, facilitating subsequent QA generation that emphasizes object dynamics rather than actions or environmental context.

\noindent\textbf{Stage 3: Grounded QA Generation.}
For each video clip corresponding to a sampled action sequence from Stage 2, we prompt Qwen3.5 \cite{qwen3.5} to generate questions, answers, and supporting temporal evidence (i.e., start and end timestamps). For each question type shown in Figure \ref{fig:qa_type}, we define task-specific system prompts to specify the requirements. The user prompt combines action annotations describing the sequence with predefined question templates, e.g., \texttt{How does the [OBJECT] change from [ACTION A] to [ACTION B]?} for Object-State Evolution questions. To represent the video content, we provide the object-centric captions obtained in Stage 2 together with a small set of sparsely sampled frames as multimodal input to Qwen3.5, supplying visual context such as scene and environment. This multimodal conditioning ensures that the generated QA pairs remain consistent with the underlying video content and aligned with the intended QA objectives.

\noindent\textbf{Stage 4: Verification and Filtering.}
Due to VLM hallucination \cite{liu2024survey}, the generated QA pairs may suffer from issues such as object-irrelevant questions, mismatched answers, misaligned evidence segments, or low overall quality. As fully manual curation is costly, we design a data cleaning pipeline that combines model-based and human verification and refinement, consisting of the following steps:
\begin{itemize}[leftmargin=0.5cm]
\vspace{-2mm}
    \item Step 1: Commonsense Filtering. We identify questions answerable without video grounding by prompting a large language model (LLM) \cite{yang2025qwen3} to generate answers using only the question text. If the text-only prediction achieves a high-quality evaluation (e.g., an LLM-as-judge score $\geq 4$, as defined in Section \ref{sec:eval_metrics}), the instance is discarded. Such questions are likely solvable via pure commonsense reasoning and would artificially inflate benchmark performance. This step filters out approximately 30\% ($\sim$4,500) of the initial QA samples.
    
    \item Step 2: Answer-Evidence Alignment. We verify alignment by inputting the question, answer, and the video segment cropped using evidence timestamps into GPT-4o mini \cite{openai2024gpt4ocard}, which outputs a binary decision indicating whether the evidence supports the answer. Accepted instances proceed to manual review, while rejected ones are directly sent for human refinement in Step 3.
    
    \item Step 3: Human Review and Refinement. We conduct human review and refinement by students with AI backgrounds majoring in computer science. First, each QA instance is independently verified twice by different student reviewers, and only those instances that pass both verifications are included in the final dataset. The others are then manually refined by correcting the questions, answers, and/or evidence. This step removes QA ambiguity, tightens evidence grounding, and improves linguistic clarity, ensuring the reliability of the benchmark.
\vspace{-2mm}
\end{itemize}

\subsection{Dataset Statistics}
\label{sec:data_statistic}

\begin{wrapfigure}{r}{0.45\textwidth}
  \vspace{-16pt}
  \centering
  \begin{subfigure}{\linewidth}
    \centering
    \includegraphics[width=\linewidth]{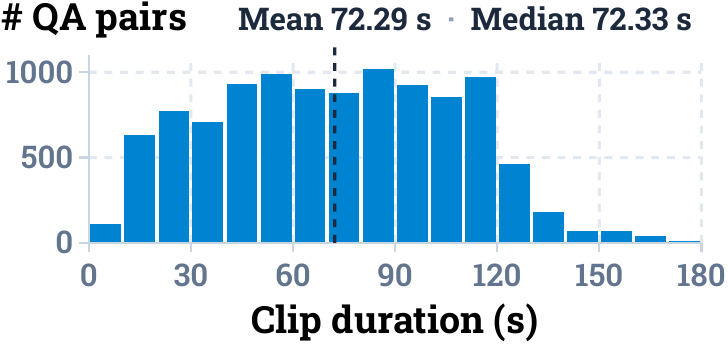}
    \caption{Video-clip length distribution.} 
    \label{fig:video_length}
  \end{subfigure}
  
  \begin{subfigure}{\linewidth}
    \centering
    \includegraphics[width=\linewidth]{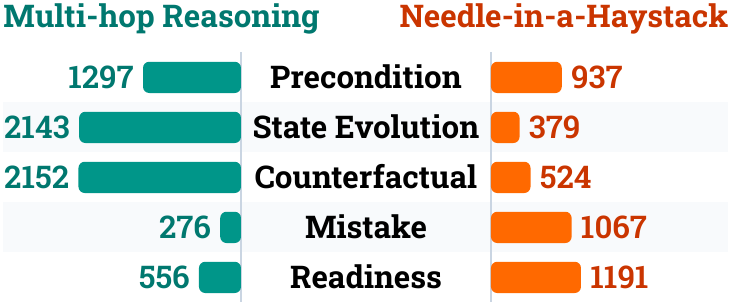}
    \caption{Reasoning categories.} 
    \label{fig:reasoning_type}
  \end{subfigure}
  
  \caption{Statistics of \textsc{ProcObject-10K}.} 
  \label{fig:object10k_stats}
\end{wrapfigure}
As shown in Figure \ref{fig:data_statistic}, after data cleaning, \textsc{ProcObject-10K} comprises 10,522 grounded QA pairs sourced from 1,799 procedural videos, covering 137 tasks across 9 domains. The dataset spans 211.28 hours of video clips in total, with an average duration of 72.29 seconds and a median of 72.33 seconds as shown in Figure \ref{fig:video_length}. We adopt a video-disjoint split with 9,472 training and 1,050 testing samples, ensuring that there is no overlap in videos. Beyond the five types of QA (Figure \ref{fig:qa_type}), we further categorize samples by temporal search patterns, as shown in Figure \ref{fig:reasoning_type}: (1) \emph{Multi-hop Reasoning}, which requires aggregating evidence across multiple or temporally separated segments, and (2) \emph{Needle-in-a-Haystack}, which involves identifying short but critical moments within long videos. Multi-hop reasoning dominates in state evolution and counterfactual reasoning, whereas mistake recognition and readiness assessment more often follow the needle-in-a-haystack pattern. 
\section{Object-centric Supervised Finetuning}
\label{sec:method}

\begin{figure}
    \centering
    \includegraphics[width=0.95\linewidth]{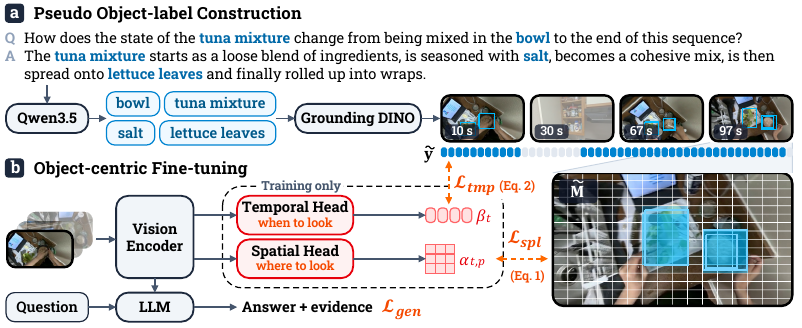}
    \caption{Overview of Object-centric SFT: (a) supporting objects extracted from QA pairs are localized to construct temporal and spatial pseudo-labels, (b) two training-only heads enforce auxiliary spatial-temporal constraints to guide \emph{where} and \emph{when} the model needs to look.}
    \vspace{-6pt}
    \label{fig:framework}
\end{figure}

Standard VideoQA fine-tuning optimizes answer generation, but does not explicitly supervise which objects and temporal states support the answer. To encourage object-centric procedural reasoning, we introduce an object-centric supervised fine-tuning framework with pseudo object-level supervision and complementary spatial-temporal constraints, as illustrated in Figure~\ref{fig:framework}. In the following, we introduce the construction of pseudo-labels in Section~\ref{sec:supervision}, the auxiliary constraints in Section~\ref{sec:constraints}, and the overall training objective in Section~\ref{sec:objective}.

\subsection{Pseudo Object-label Construction}
\label{sec:supervision}

To avoid expensive manual annotations, we construct pseudo object-labels as proxy supervision for object-centric reasoning, as shown in Figure~\ref{fig:framework}(a). For each training sample, we first use a vision-language model \cite{qwen3.5} to extract a set of supporting object phrases from the question-answer pair. These phrases correspond to the key objects required to answer the question. We then use the extracted phrases as text queries for an object detection model~\cite{liu2023grounding}, which localizes the corresponding objects in sampled video frames.

Assume that each input video includes $T$ frames, and each frame is divided into $P$ visual patches by the MLLM. For each frame, the detection model returns bounding boxes together with the associated confidence scores. We convert these detection outputs into two forms of pseudo supervision.

\textbf{Soft Patch-level Masks} $\tilde{\mathbf{M}} \in [0,1]^{T \times P}$. Each entry $\tilde m_{t,p}$ measures the overlap ratio between patch $p$ in frame $t$ and the detected object regions. A value of $0$ indicates no overlap, $1$ indicates full coverage by a bounding box, and intermediate values represent partial overlap.

\textbf{Object Presence Indicators} $\tilde{\mathbf{y}} \in \{0,1\}^{T}$. Each entry $\tilde y_t$ indicates whether frame $t$ contains the supporting objects with sufficiently high confidence. Specifically, $\tilde y_t = \mathbb{I}[\bar{s}_t > \tau]$, where $\bar{s}_t$ denotes the average confidence score of all bounding boxes in frame $t$, $\tau$ is a confidence threshold, and $\mathbb{I}[\cdot]$ is the indicator function.

Together, $\tilde{\mathbf{M}}$ provides spatial supervision over \emph{where} relevant objects appear, while $\tilde{\mathbf{y}}$ provides temporal supervision over \emph{when} they appear. To improve robustness, we further incorporate detection confidence into the loss weighting to reduce the influence of unreliable pseudo-labels during training.

\subsection{Spatial and Temporal Constraints}
\label{sec:constraints}

Based on the pseudo object-labels, we introduce two lightweight auxiliary heads on the MLLM vision encoder to encourage the model to attend to question-relevant objects and their temporal dynamics, as shown in Figure~\ref{fig:framework}(b).

\textbf{Spatial Constraint}: We introduce a lightweight spatial head after the MLLM vision encoder to predict $\alpha_{t,p} \in (0,1)$ for the $p$-th patch in frame $t$, indicating the probability that the patch overlaps with the supporting object regions. Given the pseudo-label soft mask $\tilde{\mathbf{M}}$, the spatial constraint is defined as:
\begin{equation}
\mathcal{L}_{\mathrm{spl}}
=
-\frac{1}{TP}
\sum_{t=1}^{T}\sum_{p=1}^{P}
w_t
\left[
\tilde m_{t,p}\log (\alpha_{t,p})
+
(1-\tilde m_{t,p})\log(1-\alpha_{t,p})
\right],
\end{equation}
where $w_t \in [0,1]$ is a frame-level reliability weight derived from the average confidence score of all bounding boxes in frame $t$. This constraint encourages the model to focus on spatial regions associated with supporting objects. However, spatial localization alone is insufficient for object-centric procedural reasoning, because critical object states may only appear at specific moments in the procedure. We therefore introduce a complementary temporal constraint.

\textbf{Temporal Constraint}: We introduce another lightweight temporal head after the vision encoder to predict $\beta_t \in (0,1)$ for the $t$-th frame, indicating the probability that frame $t$ contains supporting objects relevant to the question. Given the pseudo-label object presence indicator $\tilde{\mathbf{y}}$, the temporal constraint is defined as:
\begin{equation}
\mathcal{L}_{\mathrm{tmp}}
=
-\frac{1}{T}
\sum_{t=1}^{T}
\left[
\tilde y_t\log (\beta_t)
+
(1-\tilde y_t)\log(1-\beta_t)
\right].
\end{equation}
This constraint encourages the model to focus on specific frames containing key objects and their state transitions. Together, the spatial and temporal constraints encourage the model to encode not only which objects are relevant, but also where and when the relevant object states appear to support the answer.

\subsection{Training Objectives}
\label{sec:objective}

We retain the standard generative language modeling loss, $\mathcal{L}_{\mathrm{gen}}$, as the primary supervision signal for answer generation. To further encourage object-centric reasoning, we augment this objective with the spatial and temporal constraints as auxiliary losses. The final training objective is formulated as:
\begin{equation}
\mathcal{L}
=
\mathcal{L}_{\mathrm{gen}}
+
\lambda_{\mathrm{spl}}\mathcal{L}_{\mathrm{spl}}
+
\lambda_{\mathrm{tmp}}\mathcal{L}_{\mathrm{tmp}},
\end{equation}
where $\lambda_{\mathrm{spl}}$ and $\lambda_{\mathrm{tmp}}$ control the strength of the auxiliary constraints. In practice, these weights are kept small and gradually warmed up during training, allowing the auxiliary supervision to shape the visual representation without overwhelming the answer-generation objective. Notably, the auxiliary heads are only used during training to encourage the learning of object-centric visual representations. During inference, all auxiliary heads are removed to maintain efficiency.

\section{Experiments}
\label{sec:exp}

In this section, we first introduce the benchmark models and evaluation metrics in Sections \ref{sec:benchmark_models} and \ref{sec:eval_metrics}, respectively. We then present quantitative experimental results to assess the object-centric procedural reasoning capabilities of different MLLMs in Section \ref{sec:results}. Finally, we conduct ablation studies to validate the effectiveness of our proposed supervision in Section \ref{sec:ablation}, and evaluate its generalization to grounded VideoQA and embodied planning tasks in Section \ref{sec:generalization}.

\subsection{Benchmark Models}
\label{sec:benchmark_models}

To comprehensively evaluate existing models on \textsc{ProcObject-10K}, we benchmark 13 models with diverse architectures, parameter sizes, and input modalities.

\textbf{Blind LLMs:} To explore whether commonsense reasoning can solve our benchmark task, we benchmark several language-only models, including GPT-5.4-Mini \cite{openai2026gpt54mini}, Claude-Sonnet-4.6 \cite{anthropic2026claudesonnet46}, Qwen3-30B-A3B \cite{yang2025qwen3}, and Llama-3.2-3B \cite{dubey2024llama}. These models only receive textual prompts and questions without video input, forcing them to rely on internal priors to address procedural questions.

\textbf{MLLMs:} We benchmark representative commercial and open-source multimodal models that take both video and text as input. Commercial models include Gemini-3.1-Flash-Lite \cite{google2026gemini31flashlite}, GPT-5.4-Mini \cite{openai2026gpt54mini}, and Claude-Sonnet-4.6 \cite{anthropic2026claudesonnet46}. Open-source baselines include variants of the InternVL3.5 family (4B, 8B, and 38B) \cite{wang2025internvl3_5} and the Qwen3-VL family (4B, 8B, and 30B-A3B) \cite{bai2025qwen3vl}.

\textbf{Fine-tuned Model:} We also benchmark the Qwen3-VL-4B model \cite{bai2025qwen3vl} fine-tuned using object-centric SFT on \textsc{ProcObject-10K} training data to demonstrate its effectiveness.

\begin{table*}[t]
\centering
\scriptsize
\setlength{\tabcolsep}{2.5pt}
\caption{
Evaluation results on the test set of \textsc{ProcObject-10K}.
Answering is evaluated by Sentence Similarity (S.), BERT-Score F1 (B.), and LLM-as-Judge score (J.).
Grounding is evaluated by mean IoU\%, mean IoP\%, and mean IoG\%.
The \colorbox{bestgreen}{Best} and \colorbox{secondred}{Second-best} results are highlighted.}
\label{tab:main_results}
\begin{tabular}{p{2.9cm}cccccccccccccccccc}
\toprule
\multirow{3}{*}{\textbf{Methods}}
& \multicolumn{6}{c}{\textbf{Multi-hop Reasoning}}
& \multicolumn{6}{c}{\textbf{Needle-in-a-Haystack}}
& \multicolumn{6}{c}{\textbf{All}} \\
\cmidrule(lr){2-7} \cmidrule(lr){8-13} \cmidrule(lr){14-19}
& \multicolumn{3}{c}{\textbf{Answering}}
& \multicolumn{3}{c}{\textbf{Grounding}}
& \multicolumn{3}{c}{\textbf{Answering}}
& \multicolumn{3}{c}{\textbf{Grounding}}
& \multicolumn{3}{c}{\textbf{Answering}}
& \multicolumn{3}{c}{\textbf{Grounding}} \\
\cmidrule(lr){2-4} \cmidrule(lr){5-7}
\cmidrule(lr){8-10} \cmidrule(lr){11-13}
\cmidrule(lr){14-16} \cmidrule(lr){17-19}
& S. & B. & J. & IoU & IoP & IoG
& S. & B. & J. & IoU & IoP & IoG
& S. & B. & J. & IoU & IoP & IoG \\
\midrule

\multicolumn{19}{l}{\textit{\textcolor{groupgray}{Blind LLM}}} \\
GPT-5.4-Mini
& 74.0 & 89.2 & 3.13 & - & - & -
& 71.7 & 89.0 & 3.15 & - & - & -
& 73.1 & 89.1 & 3.14 & - & - & - \\

Claude-Sonnet-4.6
& 74.5 & 88.3 & 3.51 & - & - & -
& 69.2 & 87.6 & 3.08 & - & - & -
& 72.4 & 88.0 & 3.34 & - & - & - \\

Qwen3-30B-A3B
& 71.0 & 88.8 & 3.07 & - & - & -
& 68.9 & 88.7 & 2.98 & - & - & -
& 70.2 & 88.8 & 3.04 & - & - & - \\

Llama-3.2-3B
& 65.2 & 88.1 & 2.50 & - & - & -
& 60.5 & 87.9 & 2.45 & - & - & -
& 63.4 & 88.0 & 2.48 & - & - & - \\

\cdashline{1-19}[0.4pt/2pt]

\multicolumn{19}{l}{\textit{\textcolor{groupgray}{Closed-source MLLMs}}} \\
Gemini-3.1-Flash-Lite
& 68.3 & 89.7 & 3.34 & 38.7 & 71.3 & 48.2
& 75.4 & 88.9 & 2.45 & 31.0 & 46.2 & 50.8
& 71.1 & 89.4 & 2.99 & 35.7 & 61.5 & 49.2 \\

GPT-5.4-Mini
& 72.1 & 88.9 & 3.58 & 45.8 & 68.5 & 62.0
& 75.6 & 88.6 & \cellcolor{secondred}3.16 & 32.6 & 44.9 & \cellcolor{bestgreen}60.7
& 73.5 & 88.8 & 3.42 & 40.7 & 59.3 & \cellcolor{bestgreen}61.5 \\

Claude-Sonnet-4.6
& 65.4 & 85.6 & \cellcolor{bestgreen}3.86 & \cellcolor{secondred}48.1 & 70.8 & \cellcolor{bestgreen}62.6
& 70.7 & 85.6 & \cellcolor{bestgreen}3.30 & 32.5 & 46.3 & \cellcolor{secondred}55.1
& 67.5 & 85.6 & \cellcolor{bestgreen}3.64 & \cellcolor{secondred}42.0 & 61.3 & \cellcolor{secondred}59.7 \\

\cdashline{1-19}[0.4pt/2pt]

\multicolumn{19}{l}{\textit{\textcolor{groupgray}{Open-source MLLMs}}} \\
InternVL3.5-4B
& 70.8 & 88.9 & 2.30 & 19.8 & 46.4 & 32.9
& 65.4 & 88.3 & 1.86 & 12.6 & 29.8 & 32.6
& 68.7 & 88.7 & 2.13 & 17.0 & 39.9 & 32.8 \\

InternVL3.5-8B
& \cellcolor{secondred}76.6 & 90.2 & 2.88 & 22.4 & 66.5 & 27.6
& 69.4 & 89.5 & 2.29 & 16.0 & 49.1 & 21.9
& 73.8 & 89.9 & 2.65 & 19.9 & 59.7 & 25.4 \\

InternVL3.5-38B
& 76.2 & \cellcolor{secondred}90.4 & 2.84 & 27.2 & \cellcolor{secondred}72.8 & 33.1
& 70.4 & \cellcolor{secondred}89.9 & 2.39 & 22.8 & 52.8 & 30.8
& 73.9 & \cellcolor{secondred}90.2 & 2.66 & 25.5 & \cellcolor{secondred}65.0 & 32.2 \\

Qwen3-VL-4B
& 70.7 & 89.5 & 3.08 & 43.3 & 71.4 & 56.5
& 77.5 & 88.6 & 2.34 & 31.3 & 50.2 & 50.2
& 73.3 & 89.2 & 2.79 & 38.6 & 63.1 & 54.1 \\

Qwen3-VL-8B
& 71.5 & 89.8 & 3.28 & 40.9 & 72.3 & 52.8
& 78.7 & 88.6 & 2.51 & 30.3 & 50.7 & 47.1
& \cellcolor{secondred}74.3 & 89.3 & 2.98 & 36.8 & 63.9 & 50.6 \\

Qwen3-VL-30B-A3B
& 71.1 & 90.0 & 3.35 & 42.4 & 71.2 & 54.6
& \cellcolor{secondred}78.9 & 88.9 & 2.67 & \cellcolor{secondred}32.9 & \cellcolor{secondred}53.7 & 48.5
& 74.1 & 89.6 & 3.09 & 38.7 & 64.4 & 52.2 \\

\cdashline{1-19}[0.4pt/2pt]

\multicolumn{19}{l}{\textit{\textcolor{groupgray}{Object-centric Fine-tuned}}} \\
Qwen3-VL-4B
& \cellcolor{bestgreen}78.0 & \cellcolor{bestgreen}92.2 & \cellcolor{secondred}3.79 & \cellcolor{bestgreen}51.4 & \cellcolor{bestgreen}73.1 & \cellcolor{secondred}62.5
& \cellcolor{bestgreen}82.4 & \cellcolor{bestgreen}91.6 & 3.07 & \cellcolor{bestgreen}33.6 & \cellcolor{bestgreen}54.5 & 44.6
& \cellcolor{bestgreen}79.7 & \cellcolor{bestgreen}92.0 & \cellcolor{secondred}3.51 & \cellcolor{bestgreen}44.5 & \cellcolor{bestgreen}65.9 & 55.5 \\
\bottomrule
\end{tabular}
\vspace{-10pt}
\end{table*}

\begin{figure}[t]
    \centering
    \includegraphics[width=0.98\linewidth]{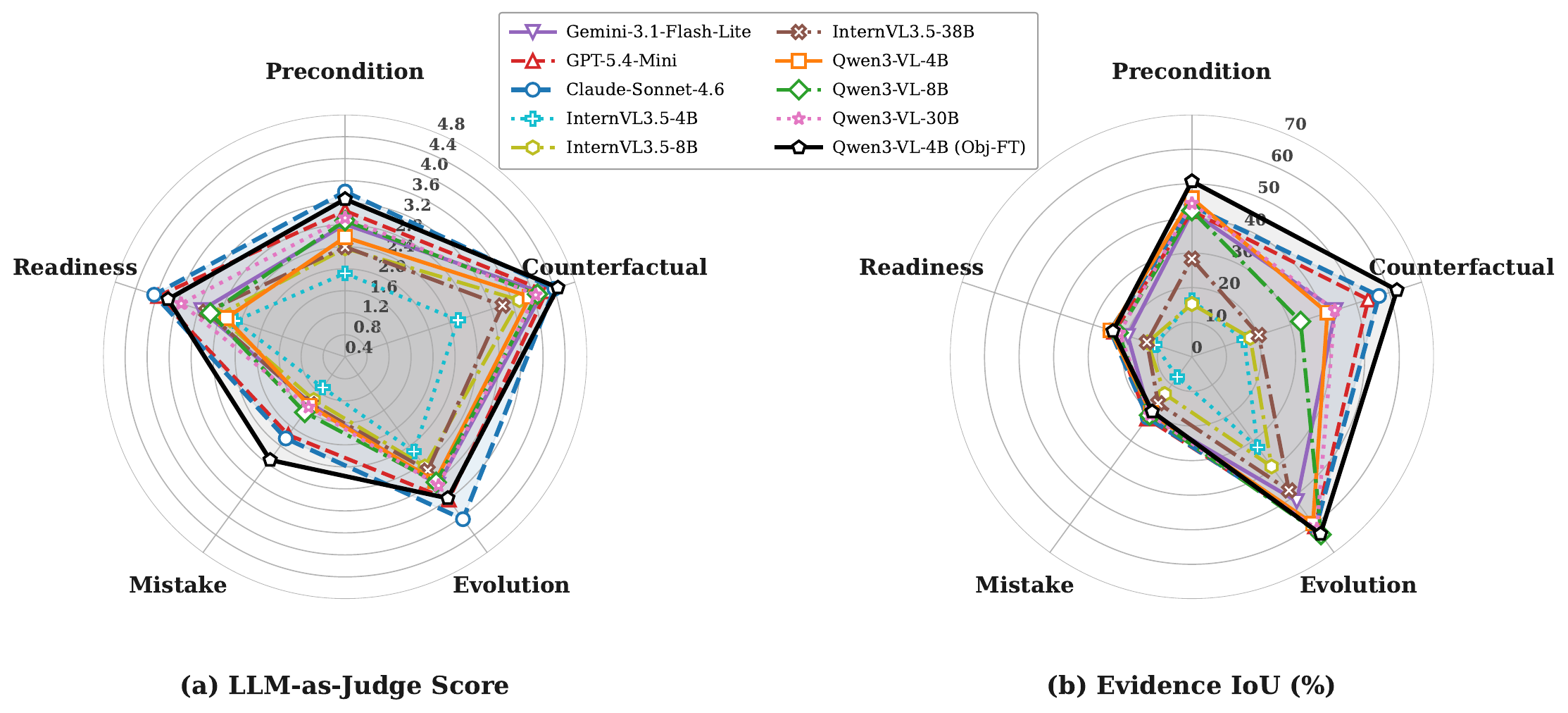}
    \caption{Performance comparison of MLLMs across five QA types on \textsc{ProcObject-10K}.}
    \vspace{-12pt}
    \label{fig:rader}
\end{figure}

\subsection{Evaluation Metrics}
\label{sec:eval_metrics}

Our benchmark includes a comprehensive evaluation that jointly measures open-ended question answering and temporal grounding accuracy, enabling rigorous assessment of model performance.

\textbf{Answering Metrics.} We evaluate the generated responses using a bi-level quantitative approach. At the textual-embedding level, we use \textit{Sentence Similarity (S.)} \cite{reimers2019sentence} and \textit{BERT-Score F1 (B.)} \cite{zhang2019bertscore} to provide reproducible measurements. At the natural-language level, we employ the LLM-as-Judge paradigm to assess linguistic coherence. Following previous works \cite{liu2025surveillancevqa, maaz2024videogpt+}, we utilize GPT-5-mini \cite{openai_gpt5_2025}, Qwen3 (4B) \cite{yang2025qwen3} and Llama-3.2 (3B) \cite{dubey2024llama} to evaluate answers independently from four dimensions using a 0–5 scoring scale: Contextual Integration for factual consistency, Detail Orientation for fine-grained completeness, Contextual Understanding for narrative and causal flow, and Temporal Understanding for event ordering and state transitions. The final \textit{LLM-as-judge score (J.)} is the average across all four dimensions and all LLM-as-judge models.

\textbf{Grounding Metrics.} To evaluate the ability to localize visual evidence, we measure the alignment between the predicted and ground-truth evidence of temporal segments. Since each QA pair in \textsc{ProcObject-10K} may be supported by multiple non-overlapping intervals, we follow previous work \cite{chen2025grounded} and adopt a set-level Intersection-over-Union (IoU) metric. For each sample with $m$ predicted spans $\hat{\mathcal{T}} = \{\hat{T}_i\}_{i=1}^{m}$ and $n$ ground-truth spans $\mathcal{T} = \{T_j\}_{j=1}^{n}$:
\begin{equation}
    \text{IoU}(\mathcal{T}, \hat{\mathcal{T}}) = \frac{\sum_{i=1}^{m}\sum_{j=1}^{n} |\hat{T}_i \cap T_j|}{\left| \bigcup_{i=1}^{m} \hat{T}_i \cup \bigcup_{j=1}^{n} T_j \right|}.
\end{equation}
In the following experiments, we report the mean \textit{IoU} across all samples. We additionally report mean \textit{IoP} and mean \textit{IoG} \cite{chen2025grounded}, which replace the denominator with $\left|\bigcup_{i=1}^{m} \hat{T}_i\right|$ and $\left|\bigcup_{j=1}^{n} T_j\right|$ respectively, serving as precision- and recall-style counterparts.

\subsection{Quantitative Results and Analysis}
\label{sec:results}

Table \ref{tab:main_results} and Figure \ref{fig:rader} demonstrate the experimental results on \textsc{\textsc{ProcObject-10K}}. We have the following findings: \textbf{1) Overall grounding performance remains constrained:} Across all benchmark models, temporal grounding precision is notably low, with mIoU scores consistently falling below 45\%. \textbf{2) A sharp answering-grounding gap reveals reliance on priors:} There is a significant discrepancy between answering plausibility and grounding precision, suggesting that models often rely on internal linguistic and commonsense priors rather than actual visual evidence. \textbf{3) Blind LLMs show competitive linguistic performance:} The reliance on commonsense reasoning is further validated by language-only models, such as GPT-5.4-Mini (Blind), which achieves an Answering Judge score of 3.14 despite having no video input, only marginally lower than its MLLM counterpart's 3.42. \textbf{4) Parameter scaling yields limited gains:} Increasing model size does not inherently solve object-centric understanding, as observed by the marginal improvements when scaling from 4B to greater than 30B variants of Qwen3-VL and InternVL3.5 families. \textbf{5) ``Needle-in-a-Haystack'' reasoning is more challenging:} Pinpointing very short-term but critical object-state appearance or transitions within long sequences is harder than \textit{Multi-hop Reasoning}, reflecting the difficulty of identifying sparse dynamics within complex human-object interactions. \textbf{6) Object-centric SFT demonstrates superior effectiveness:} Our fine-tuned Qwen3-VL-4B model achieves the highest performance across most metrics, with its 4B-parameter architecture surpassing much larger open-source and proprietary baselines, illustrating the effectiveness of our finetuning paradigm. Detailed analysis of performance improvement will be conducted through ablation study in Section \ref{sec:ablation}. 

\begin{table*}[t]
\centering
\renewcommand{\arraystretch}{1.1}
\begin{minipage}[t]{0.6\textwidth}
\centering
\caption{
Evaluation on \textsc{MultiHop-EgoQA} \cite{chen2025grounded}.}
\label{tab:multihop_egoqa}
\resizebox{\linewidth}{!}{%
\begin{tabular}{@{} p{3.6cm}ccccccc } 
\toprule
\multirow{2}{*}{\textbf{Methods}}
& \multirow{2}{*}{\textbf{Input}}
& \multicolumn{4}{c}{\textbf{Temporal Grounding}}
& \multicolumn{2}{c}{\textbf{Question Answering}} \\
\cmidrule(lr){3-6} \cmidrule(lr){7-8}
& 
& \textbf{mIoP} & \textbf{mIoG} & \textbf{IoU@0.3} & \textbf{mIoU}
& \textbf{Sent. Sim.} & \textbf{LLM Score} \\
\midrule
Human
& --
& 71.8 & 81.0 & 87.0 & 61.8
& 74.3 & 7.5 \\

\noalign{\vspace{1pt}}
\cdashline{1-8}[0.4pt/2pt]
\noalign{\vspace{1pt}}

\multicolumn{8}{@{}l}{\textit{\textcolor{groupgray}{Zero-shot Inference}}} \\
GPT-4o
& 60 f
& 18.9 & 24.4 & 12.0 & 12.2
& 73.7 & \cellcolor{bestgreen}5.4 \\

InternVL2-8B
& 30 f
& 11.8 & 24.0 & 6.3 & 6.6
& 71.9 & 4.5 \\

LLaVA-NeXT-Video-7B
& 32 f
& -- & -- & -- & --
& 62.1 & 4.2 \\

TimeChat-7B
& 96 f
& 10.2 & 5.6 & 3.0 & 3.6
& 58.9 & 3.3 \\

VTimeLLM-7B
& 100 f
& 12.4 & 28.2 & 8.8 & 9.2
& 70.5 & 4.3 \\

Qwen3-VL-4B
& 60 f
& 19.0 & \cellcolor{secondred}38.1 & 15.5 & 13.1
& 50.0 & 3.2 \\

\noalign{\vspace{1pt}}
\cdashline{1-8}[0.4pt/2pt]
\noalign{\vspace{1pt}}

\multicolumn{8}{@{}l}{\textit{\textcolor{groupgray}{Fine-tuned on \textsc{MultiHop-EgoQA}}}} \\
GeLM-7B
& 180 f
& 23.7 & \cellcolor{bestgreen}41.0 & 18.2 & 16.7
& \cellcolor{bestgreen}75.0 & 4.8 \\

\noalign{\vspace{1pt}}
\cdashline{1-8}[0.4pt/2pt]
\noalign{\vspace{1pt}}

\multicolumn{8}{@{}l}{\textit{\textcolor{groupgray}{Fine-tuned on \textsc{\textsc{ProcObject-10K}}}}} \\
Qwen3-VL-4B$^\ddagger$
& 60 f
& \cellcolor{secondred}31.4 & 32.1 & \cellcolor{secondred}25.2 & \cellcolor{secondred}19.8
& 73.6 & \cellcolor{secondred}5.3 \\

Qwen3-VL-4B
& 60 f
& \cellcolor{bestgreen}34.8 & 37.3 & \cellcolor{bestgreen}30.8 & \cellcolor{bestgreen}22.6
& \cellcolor{secondred}74.6 & \cellcolor{bestgreen}5.4 \\

\bottomrule
\end{tabular}%
}
\end{minipage}%
\hfill
\begin{minipage}[t]{0.38\textwidth}
\centering
\caption{
Evaluation on ALFRED \cite{ALFRED20}. 
}
\label{tab:flare_qwen_results}
\resizebox{\linewidth}{!}{%
\begin{tabular}{@{} lcccc }
\toprule
\multirow{2}{*}{\textbf{Methods}}
& \multicolumn{2}{c}{\textbf{Test Seen}}
& \multicolumn{2}{c}{\textbf{Test Unseen}} \\
\cmidrule(lr){2-3} \cmidrule(lr){4-5}
& \textbf{SR} & \textbf{GC}
& \textbf{SR} & \textbf{GC} \\
\midrule

\multicolumn{5}{@{}l}{\textit{\textcolor{groupgray}{Zero-shot MLLM}}} \\

GPT-4
& 37.40 & \cellcolor{secondred}48.30 & 34.90 & 44.90 \\

Qwen3-VL-4B
& 30.50 & 43.60 & 29.40 & 40.50 \\

\noalign{\vspace{1pt}}
\cdashline{1-5}[0.4pt/2pt]
\noalign{\vspace{1pt}}

\multicolumn{5}{@{}l}{\textit{\textcolor{groupgray}{FLARE+Zero-shot MLLM Planner}}} \\

FLARE+LLaMA2
& 16.96 & 24.84 & 17.79 & 27.40 \\

FLARE+Vicuna
& 20.61  & 29.57  & 22.04  & 33.57 \\

FLARE+GPT-3.5
& 32.55  & 42.02  & 31.79  & 43.94 \\

FLARE+GPT-4
& \cellcolor{secondred}40.05  & \cellcolor{bestgreen}48.84 & \cellcolor{secondred}40.88  & \cellcolor{secondred}51.72 \\

FLARE+Qwen3-VL-4B
& 31.50 & 44.20 & 32.01 & 44.62 \\

\noalign{\vspace{1pt}}
\cdashline{1-5}[0.4pt/2pt]
\noalign{\vspace{1pt}}

\multicolumn{5}{@{}l}{\textit{\textcolor{groupgray}{FLARE+Fine-tuned MLLM Planner}}} \\

FLARE+Qwen3-VL-4B$^\ddagger$
& 36.97 & 46.74 & 38.02 & 50.08 \\

FLARE+Qwen3-VL-4B
& \cellcolor{bestgreen}40.50 & 48.20 & \cellcolor{bestgreen}41.33 & \cellcolor{bestgreen}51.97 \\

\bottomrule
\end{tabular}%
}
\end{minipage}
\end{table*}

\begin{figure}[t]
    \centering
    \begin{subfigure}[b]{0.485\textwidth}
        \centering
        \includegraphics[width=\textwidth]{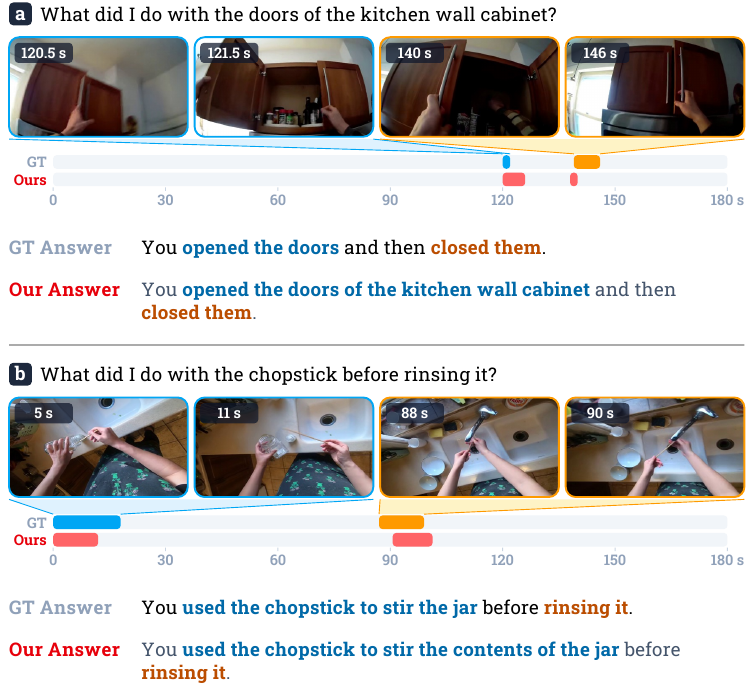}
        \caption{Generic grounded video question-answering.}
        \label{fig:gen_vqa}
    \end{subfigure}
    \hfill 
    \begin{subfigure}[b]{0.49\textwidth}
        \centering
        \includegraphics[width=\textwidth]{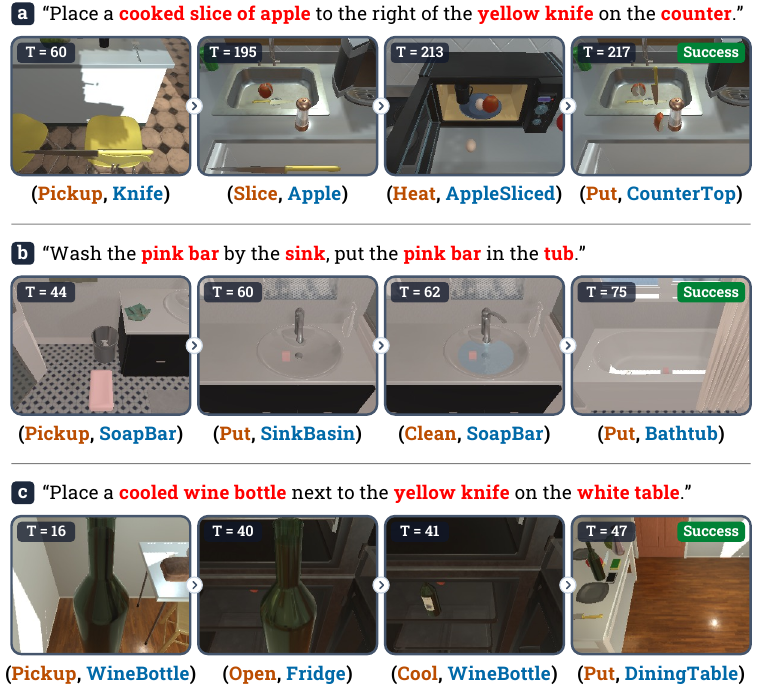}
        \caption{Embodied planning.}
        \label{fig:gen_planning}
    \end{subfigure}
    
    \caption{Visualization of zero-shot generalization of learned object-centric priors from \textsc{ProcObject-10K} on (a) generic grounded VideoQA \cite{chen2025grounded} and (b) long-horizon embodied planning \cite{ALFRED20} tasks.}
    \vspace{-6pt}
    \label{fig:generalization}
\end{figure}

\subsection{Ablation Study}
\label{sec:ablation}

\begin{wraptable}{r}{0.32\textwidth}
\vspace{-31pt}
\centering
\small
\setlength{\tabcolsep}{4pt} 
\caption{Ablation study results.}
\label{tab:ablation}
\begin{tabular}{ccccc}
\toprule
$\mathcal{L}_{\text{gen}}$ & $\mathcal{L}_{\text{spl}}$ & $\mathcal{L}_{\text{tmp}}$ & J. & IoU \\
\midrule
\checkmark &            &            & 3.18 & 42.4 \\
\checkmark & \checkmark &            & 3.21 & 42.5 \\
\checkmark &            & \checkmark & 3.26 & 43.4 \\
\checkmark & \checkmark & \checkmark & \textbf{3.51} & \textbf{44.5} \\
\bottomrule
\end{tabular}
\end{wraptable}

We conduct an ablation study on \textsc{ProcObject-10K} to evaluate the effectiveness of each training objective. Table \ref{tab:ablation} shows that all losses contribute positively to performance, with the temporal constraint providing more substantial gains. The full loss achieves the best results, validating the effectiveness of combining spatial and temporal supervision to model the dynamic evolution of objects.

\subsection{Generalization Analysis}
\label{sec:generalization}

To evaluate the generalization of object-centric procedural priors learned from \textsc{ProcObject-10K}, we test the fine-tuned models on (1) a zero-shot grounded VideoQA task using the \textsc{MultiHop-EgoQA} benchmark~\cite{chen2025grounded}, and (2) a zero-shot embodied planning task using the ALFRED benchmark~\cite{ALFRED20}. Visualization and quantitative results are shown in Figure \ref{fig:generalization} and Tables \ref{tab:multihop_egoqa}--\ref{tab:flare_qwen_results}, respectively. Qwen3-VL-4B$^\ddagger$ denotes the model variant fine-tuned without incorporating spatial-temporal constraints.

\textbf{Grounded VideoQA.}
The \textsc{MultiHop-EgoQA} dataset~\cite{chen2025grounded} focuses on instructional videos, but evaluates generic multi-hop procedural questions rather than strictly object-centric reasoning tasks. We follow its official LLM-as-Judge protocol for a fair comparison, where scores range from 0 to 10 (higher is better). As shown in Table~\ref{tab:multihop_egoqa}, even when trained using only the standard generative objective, Qwen3-VL-4B$^\ddagger$ substantially outperforms the zero-shot baseline and achieves performance surpassing GeLM-7B, which is specifically fine-tuned on the \textsc{MultiHop-EgoQA} training set. This result indicates that \textsc{ProcObject-10K} enables models to learn transferable object-centric procedural representations that generalize effectively to broader video understanding tasks. Furthermore, the model trained with our full object-centric SFT framework (Qwen3-VL-4B) yields additional gains, achieving the best zero-shot performance across both grounding (22.6\% mIoU and 30.8\% IoU@0.3) and answering (5.4 LLM Score) metrics. As visualized in Figure~\ref{fig:gen_vqa}, our model accurately localizes disjoint temporal evidence intervals across long horizons (e.g., sequentially opening and closing cabinet doors, or stirring a jar with a chopstick prior to rinsing it) and generates precise answers aligned with the ground truth.

\textbf{Embodied Planning.}
ALFRED~\cite{ALFRED20} is a benchmark requiring embodied agents to execute long-horizon household tasks from natural language instructions across both seen and unseen test environments. We evaluate transfer performance by integrating models fine-tuned on \textsc{ProcObject-10K} as zero-shot planners into the FLARE framework~\cite{kim2025multimodal}, a prior method specifically designed for this benchmark. Table~\ref{tab:flare_qwen_results} shows that our fine-tuned Qwen3-VL-4B variants consistently improve over the zero-shot baseline. Even when trained only with the standard generative objective, FLARE+Qwen3-VL-4B$^\ddagger$ outperforms several larger zero-shot MLLM planners. Incorporating our full spatial-temporal supervision, FLARE+Qwen3-VL-4B achieves the best overall results (40.50\% SR on Test Seen and 41.33\% SR on Test Unseen), even surpassing the proprietary FLARE+GPT-4 baseline. Qualitative examples in Figure~\ref{fig:gen_planning} further illustrate how learned object-state priors enable the agent to decompose complex instructions (e.g., slicing and heating an apple, or washing a soap bar before placing it in the tub) into valid, precondition-aware action sequences, highlighting the value of \textsc{ProcObject-10K} for downstream embodied reasoning.
\section{Conclusion and Discussion}
We present \textsc{ProcObject-10K}, a benchmark that reframes procedural video understanding around object state evolution rather than action sequences. Evaluation results on leading MLLMs reveal a persistent answering-grounding gap: models generate plausible answers while failing to localize the supporting evidence, indicating reliance on linguistic priors over fine-grained object dynamics. As a step toward closing this gap, we further provide an object-centric SFT baseline that narrows the gap and yields representations transferable to other grounded VideoQA and embodied planning tasks. We hope \textsc{ProcObject-10K} could catalyze further research on object-centric procedural reasoning with broader domain coverage and tighter integration with embodied agents and world models.

\begin{ack}
This research was partially sponsored by the MSU Jenison Fund and Army Research Office (under Grant Number W911NF-24-1-0385). The views and conclusions contained in this document are those of the authors and should not be interpreted as representing the official policies, either expressed or implied, of the Army Research Office or the U.S. Government. The U.S. Government is authorized to reproduce and distribute reprints for Government purposes notwithstanding any copyright notation herein.
\end{ack}

{
\small
\bibliographystyle{plain}
\bibliography{neurips_2026}
}

\clearpage
\appendix

\begin{center}
{\Large\bfseries Supplementary Material}
\end{center}

\noindent This supplementary material is organized as follows:
\begin{itemize}[leftmargin=0.5cm,itemsep=1pt,topsep=2pt]
    \item Appendix~\ref{app:benchmark}: additional details of \textsc{ProcObject-10K}, including data verification, dataset composition, evidence statistics, and qualitative examples.
    \item Appendix~\ref{app:sft_details}: implementation details of object-centric SFT, including the training data format, pseudo-label construction, hyperparameters, and compute resources.
    \item Appendix~\ref{app:limitations}: limitations and future work.
    \item Appendix~\ref{app:reproducibility}: reproducibility and release notes.
    \item Appendix~\ref{app:llm_usage}: statement on the usage of large language models (LLMs).
    \item Appendix~\ref{app:prompts}: prompt templates for QA generation and evaluation.
\end{itemize}

\section{Additional Details of \textsc{ProcObject-10K}}
\label{app:benchmark}

This section complements Section~\ref{sec:dataset}. Throughout the supplementary material, the five QA types (Precondition Grounding, Object-State Evolution, Counterfactual Reasoning, Mistake Recognition, and Readiness Assessment) are abbreviated as Precondition, State Evolution, Counterfactual, Mistake, and Readiness in tables and figures.

\subsection{Data Verification and Human Review}
\label{app:data_cleaning}

\begin{figure}[t]
    \centering
    \includegraphics[width=0.9\linewidth]{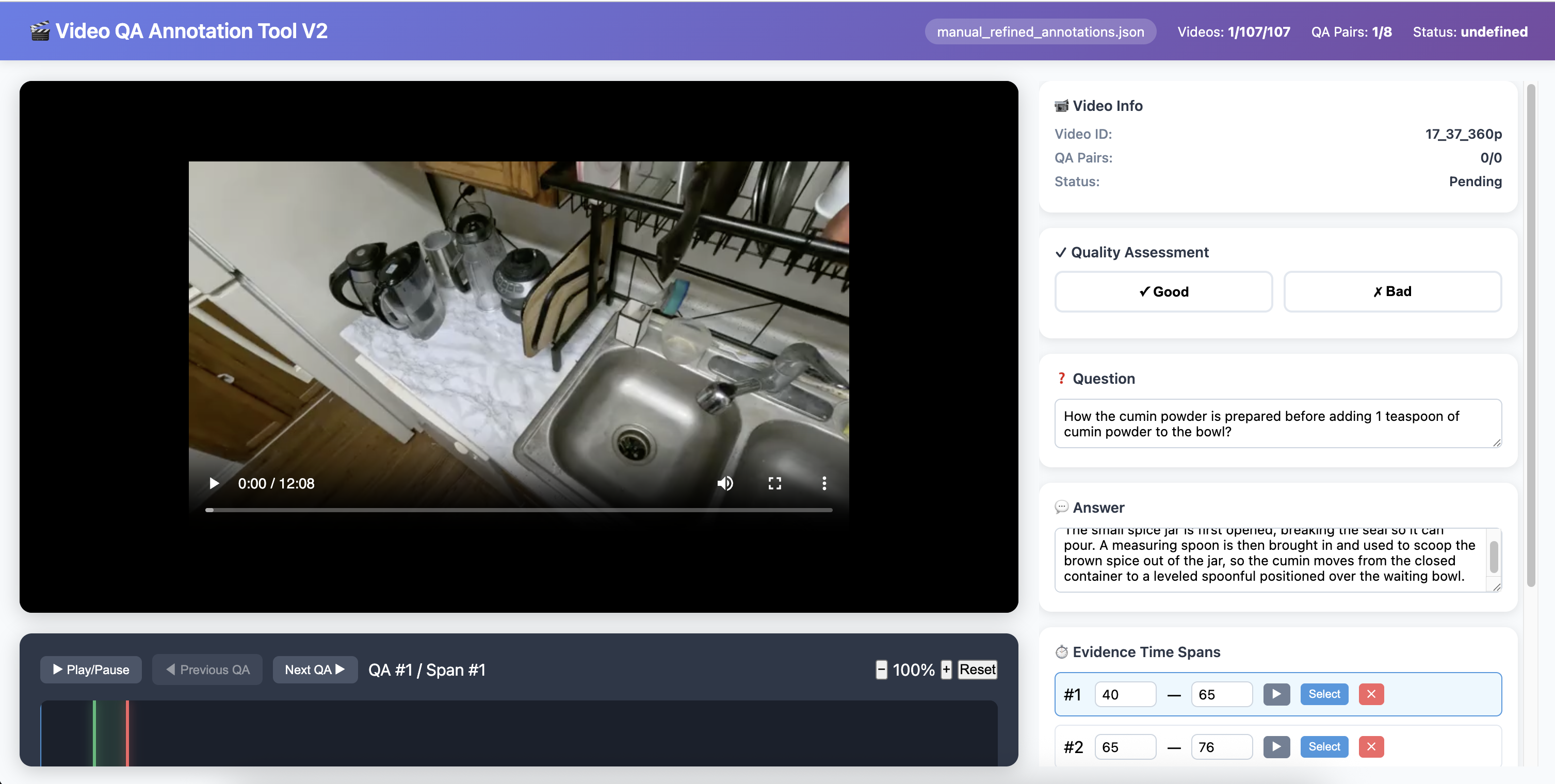}
    \caption{Web-based annotation tool for manual QA review and refinement. Reviewers play the clip, mark each QA pair as good or bad, and edit its question, answer, and evidence spans.}
    \label{fig:webpage}
\end{figure}

This subsection details Stage~4 of the data generation pipeline (Section~\ref{sec:data_generate}). Stages~1--3 produce 15,090 candidate QA pairs, which are cleaned in three steps:
\begin{itemize}[leftmargin=0.5cm,itemsep=1pt,topsep=2pt]
    \item \textbf{Commonsense filtering.} A text-only LLM~\cite{yang2025qwen3} answers each question without access to the video. An instance is discarded if this answer achieves an LLM-as-judge score $\geq 4$, since such questions can be solved by commonsense alone. This step removes approximately 30\% ($\sim$4,500) of the candidates.
    \item \textbf{Answer--evidence alignment.} GPT-4o mini~\cite{openai2024gpt4ocard} receives the question, the answer, and the video segment cropped by the evidence timestamps, and outputs a binary decision on whether the evidence supports the answer. Accepted instances proceed to human review, while rejected ones are sent directly to human refinement.
    \item \textbf{Human review and refinement.} Each instance is independently verified by two student reviewers with AI backgrounds majoring in computer science, who check four criteria: (1) the mentioned object is visible in or implied by the video segment, (2) the answer is consistent with the video rather than with commonsense alone, (3) the evidence spans support the answer, and (4) the language is unambiguous. Instances that pass both reviews are accepted; the others are refined by correcting the question, answer, and/or evidence, or removed if they cannot be repaired.
\end{itemize}
The final benchmark contains 10,522 QA pairs grounded in 1,799 videos. To facilitate review and refinement, we built a local web-based annotation tool, shown in Figure~\ref{fig:webpage}.

\subsection{Dataset Composition and Splits}
\label{app:dataset_composition}

\begin{figure}[t]
    \centering
    \includegraphics[width=\linewidth]{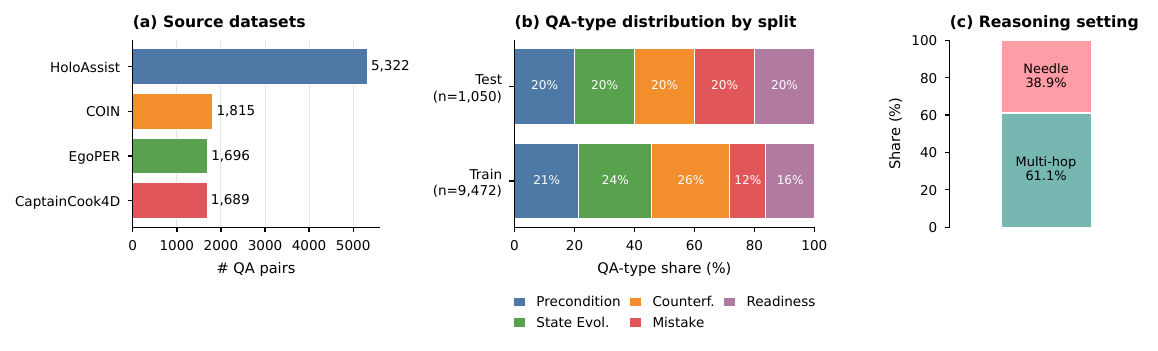}
    \caption{Dataset breakdown. (a) Number of QA pairs contributed by each source dataset. (b) QA-type distribution of the training and test splits: the test set is balanced across QA types, while the training set preserves the distribution remaining after filtering. (c) Proportions of Multi-hop Reasoning and Needle-in-a-Haystack samples.}
    \label{fig:app_dataset_breakdown}
\end{figure}

Table~\ref{tab:domain_stats} lists the per-domain statistics underlying Figure~\ref{fig:data_statistic}. \textsc{ProcObject-10K} covers 137 tasks across 9 domains, with Assembly \& Technology and Cooking \& Food as the two largest domains (86.9\% of all QA pairs). The QA pairs are curated from four source datasets (Figure~\ref{fig:app_dataset_breakdown}(a)), which provide both egocentric (CaptainCook4D, EgoPER, and HoloAssist) and exocentric (COIN) views. Following the reasoning categories in Section~\ref{sec:data_statistic}, 61.1\% of the QA pairs require Multi-hop Reasoning and 38.9\% follow the Needle-in-a-Haystack pattern (Figure~\ref{fig:app_dataset_breakdown}(c); see Figure~\ref{fig:reasoning_type} for the breakdown by QA type).

We adopt a video-disjoint train/test split, so that no video appears in both splits (Table~\ref{tab:app_dataset_summary}). The training set preserves the QA-type distribution remaining after filtering, whereas the test set is balanced with exactly 210 QA pairs per QA type (Figure~\ref{fig:app_dataset_breakdown}(b)), enabling a fair comparison across all reasoning dimensions.

\begin{table}[t]
    \centering
    \small
    \setlength{\tabcolsep}{4pt}
    \begin{minipage}[t]{0.53\linewidth}
        \centering
        \caption{Domain-level statistics.}
        \label{tab:domain_stats}
        \begin{tabular}{@{}lrrr@{}}
            \toprule
            Domain & QA pairs & Videos & Tasks \\
            \midrule
            Assembly \& Technology  & 5,322 & 792 & 9 \\
            Cooking \& Food         & 3,823 & 561 & 54 \\
            Gadgets                 & 398   & 123 & 18 \\
            Leisure \& Crafts       & 284   & 84  & 14 \\
            Electrical Appliance    & 252   & 82  & 12 \\
            Science \& Craft        & 190   & 70  & 13 \\
            Fruit \& Plant Care     & 107   & 37  & 5 \\
            Housework               & 101   & 29  & 4 \\
            Nursing \& Care         & 45    & 21  & 8 \\
            \midrule
            Total                   & 10,522 & 1,799 & 137 \\
            \bottomrule
        \end{tabular}
    \end{minipage}%
    \hfill
    \begin{minipage}[t]{0.44\linewidth}
        \centering
        \caption{Video-disjoint train/test split.}
        \label{tab:app_dataset_summary}
        \begin{tabular}{@{}lrrr@{}}
            \toprule
             & Train & Test & All \\
            \midrule
            Videos          & 1,645 & 154   & 1,799 \\
            QA pairs        & 9,472 & 1,050 & 10,522 \\
            \midrule
            Precondition    & 2,024 & 210 & 2,234 \\
            State Evolution & 2,312 & 210 & 2,522 \\
            Counterfactual  & 2,466 & 210 & 2,676 \\
            Mistake         & 1,133 & 210 & 1,343 \\
            Readiness       & 1,537 & 210 & 1,747 \\
            \bottomrule
        \end{tabular}
    \end{minipage}
\end{table}

\subsection{Evidence and Language Statistics}
\label{app:evidence_stats}

\begin{figure}[t]
    \centering
    \includegraphics[width=\linewidth]{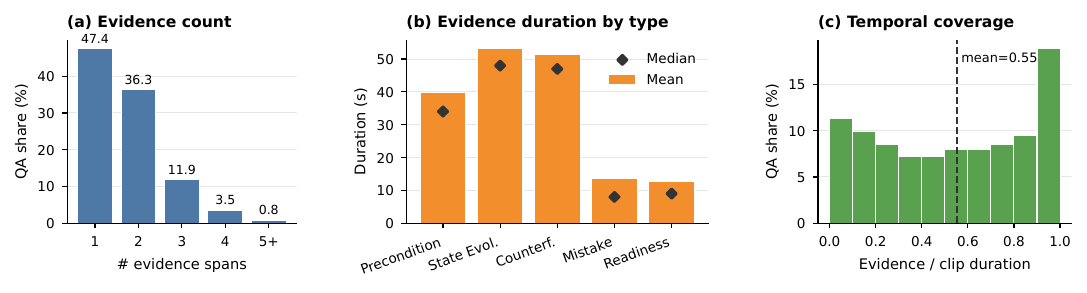}
    \caption{Evidence statistics. (a) Distribution of the number of evidence spans per question. (b) Mean and median duration of the evidence union per QA type. (c) Ratio of evidence duration to clip duration. Most questions require one or two evidence spans, and for most samples the evidence is neither a single moment nor the entire clip, making temporal localization meaningful and challenging.}
    \label{fig:app_evidence_stats}
\end{figure}

The sampled video clips last 72.29\,s on average, with a median of 72.33\,s (Figure~\ref{fig:video_length}). Each question is supported by 1.74 evidence spans on average, with a median of 2 and a maximum of 11 spans. The union of the evidence spans of a question covers 38.13\,s on average, corresponding to 55.2\% of the clip duration (Figure~\ref{fig:app_evidence_stats}). Table~\ref{tab:app_evidence_by_type} breaks these statistics down by QA type: Mistake and Readiness questions are mostly supported by a single short span, in line with their predominantly Needle-in-a-Haystack pattern, whereas State Evolution and Counterfactual questions involve about two spans and more than 50\,s of evidence. Questions and answers contain 17.19 and 31.31 words on average, respectively, indicating that the benchmark is not reducible to short keyword matching.

\begin{table}[t]
    \centering
    \small
    \caption{Evidence statistics by QA type. Durations are computed over the union of the evidence spans of each question.}
    \label{tab:app_evidence_by_type}
    \begin{tabular}{@{}lrccc@{}}
        \toprule
        QA type & QA pairs & Avg.\ spans & Median spans & Avg.\ evidence duration \\
        \midrule
        Precondition    & 2,234 & 1.63 & 1 & 39.89\,s \\
        State Evolution & 2,522 & 2.19 & 2 & 53.10\,s \\
        Counterfactual  & 2,676 & 1.96 & 2 & 51.54\,s \\
        Mistake         & 1,343 & 1.20 & 1 & 13.47\,s \\
        Readiness       & 1,747 & 1.32 & 1 & 12.66\,s \\
        \midrule
        All             & 10,522 & 1.74 & 2 & 38.13\,s \\
        \bottomrule
    \end{tabular}
\end{table}

\subsection{Qualitative Examples}
\label{app:qualitative}

\begin{figure}[tp]
    \centering
    \includegraphics[width=\linewidth]{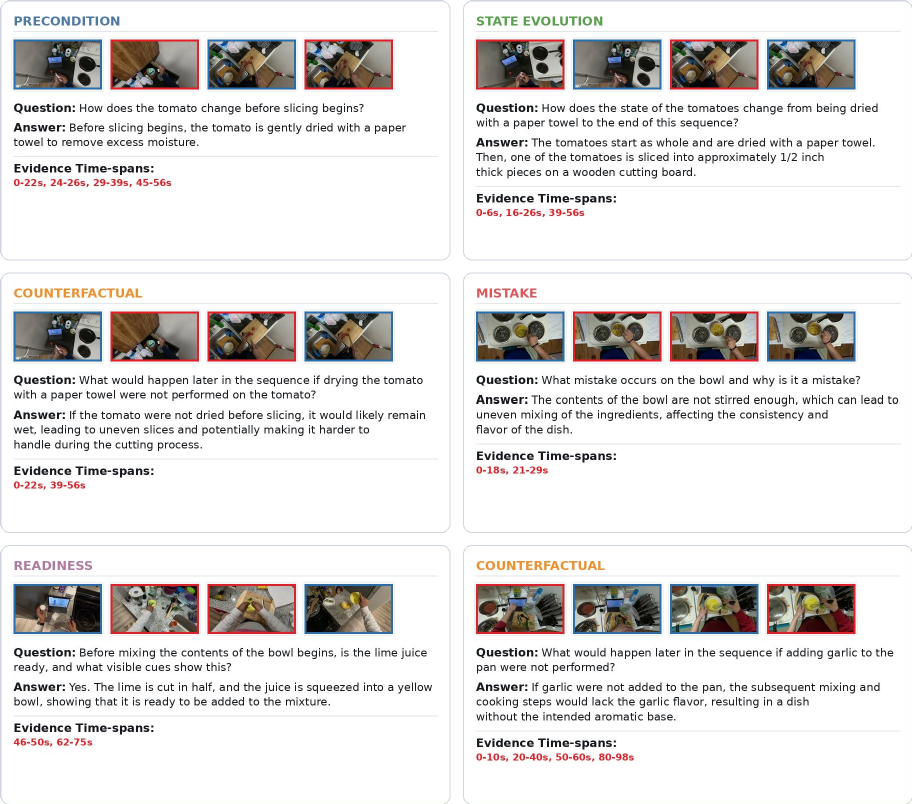}
    \caption{Examples from \textsc{ProcObject-10K} (set 1). Evidence spans are shown in red; red-bordered frames lie inside the evidence spans and blue-bordered frames lie outside.}
    \label{fig:app_qa_examples_set1}
\end{figure}

\begin{figure}[tp]
    \centering
    \includegraphics[width=\linewidth]{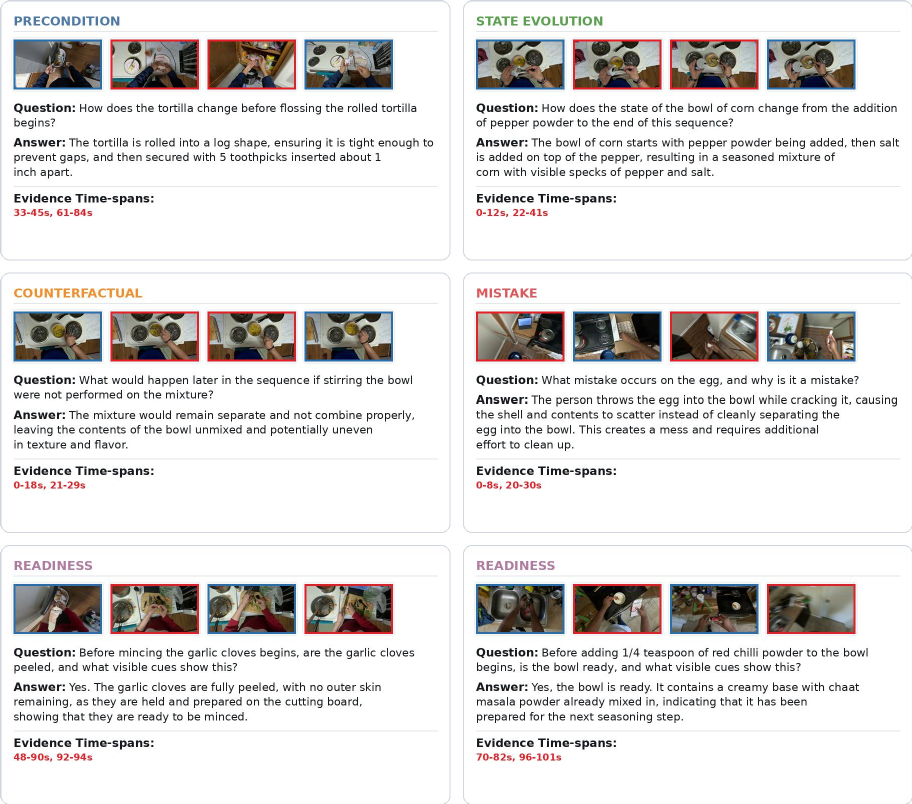}
    \caption{Examples from \textsc{ProcObject-10K} (set 2). Evidence spans are shown in red; red-bordered frames lie inside the evidence spans and blue-bordered frames lie outside.}
    \label{fig:app_qa_examples_set2}
\end{figure}

Figures~\ref{fig:app_qa_examples_set1} and~\ref{fig:app_qa_examples_set2} present additional examples covering all five QA types. Each card shows four frames sampled from the clip, the open-ended question and answer, and the ground-truth evidence spans. Frames with red borders lie inside the evidence spans, while frames with blue borders provide context outside the evidence. The answers are grounded in observable object states rather than action names alone.

\section{Implementation Details of Object-Centric SFT}
\label{app:sft_details}

This section complements Section~\ref{sec:method}. Training proceeds in two stages that share the same LoRA adapter. \emph{Standard SFT} first optimizes only the generative loss $\mathcal{L}_{\mathrm{gen}}$ on the answer and evidence. \emph{Object-centric SFT} then continues from the standard SFT adapter and optimizes the full objective $\mathcal{L}=\mathcal{L}_{\mathrm{gen}}+\lambda_{\mathrm{spl}}\mathcal{L}_{\mathrm{spl}}+\lambda_{\mathrm{tmp}}\mathcal{L}_{\mathrm{tmp}}$ (Section~\ref{sec:objective}).

\subsection{Training Data Format}
\label{app:sft_format}

The 9,472 training QA pairs are converted into a video-instruction format. Each sample contains one user turn with the video and the question, and one assistant turn whose target is a JSON response containing the answer and the evidence timestamps. We hold out 5\% of the training QA pairs (474 samples) for validation and train on the remaining 8,998 samples. Videos are encoded with the Qwen3-VL chat template and sampled at 2\,FPS with at most $T{=}48$ frames; for standard SFT, each temporal patch of two consecutive frames is limited to 1,024 visual tokens. The maximum text--video sequence length is 16,384 tokens.

\subsection{Pseudo Object Supervision}
\label{app:pseudo_object_supervision}

Pseudo labels (Section~\ref{sec:supervision}) are generated only for the training split. For each sample, a text-only Qwen3.5 model (Qwen3.5-35B-A3B) extracts up to five concrete supporting object phrases from the QA pair (question, answer, and evidence). The extractor is constrained to return visible noun phrases only, excluding people, actions, attributes, and abstract concepts. GroundingDINO~\cite{liu2023grounding} then uses these phrases as text queries to localize the objects in the same $T{=}48$ frames that are fed to the MLLM. The detections are converted into (1) the soft patch-level masks $\tilde{\mathbf{M}}$ over a $16\times16$ grid per frame and (2) the frame-level object presence indicators $\tilde{\mathbf{y}}$. Detection confidences are retained as reliability weights, so that low-confidence pseudo labels contribute less to the auxiliary losses: the per-frame weight $w_t$ in $\mathcal{L}_{\mathrm{spl}}$ is derived from the average box confidence of frame $t$, and a sample-level confidence further scales both auxiliary losses. Table~\ref{tab:app_pseudo_label_settings} summarizes the settings.

\begin{table}[!ht]
    \centering
    \small
    \caption{Pseudo-label construction settings for object-centric SFT.}
    \label{tab:app_pseudo_label_settings}
    \begin{tabular}{@{}>{\raggedright\arraybackslash}p{4.2cm}>{\raggedright\arraybackslash}p{8.6cm}@{}}
        \toprule
        Component & Setting \\
        \midrule
        Object phrase extractor & Qwen3.5-35B-A3B (text-only), at most 5 phrases per sample \\
        Object detector & GroundingDINO-base \\
        Box / text threshold & 0.30 / 0.25 \\
        Supervised frames & $T{=}48$, identical to the MLLM video input (2\,FPS, at most 48 frames) \\
        Spatial label $\tilde{\mathbf{M}}$ & Soft box--patch overlap on a $16\times16$ grid per frame \\
        Confidence weighting & Per-frame weight $w_t$ in $\mathcal{L}_{\mathrm{spl}}$; sample-level weight on both auxiliary losses \\
        \bottomrule
    \end{tabular}
\end{table}

\subsection{Optimization Details}
\label{app:optimization}

We use Qwen3-VL-4B-Instruct~\cite{bai2025qwen3vl} as the base MLLM and train LoRA adapters with rank 16 and $\alpha{=}32$. Standard SFT applies LoRA to all linear layers. Object-centric SFT jointly trains the language LoRA, a vision LoRA on the top third of the vision encoder layers, the multimodal aligner, and the two lightweight auxiliary heads, each with its own learning rate. The auxiliary loss weights are warmed up from zero over the first 10\% of training steps and decayed to 50\% of their base values over the final 25\%. The auxiliary heads are used only during training and are removed at inference time. Table~\ref{tab:app_training_hparams} lists all hyperparameters.

\begin{table}[t]
    \centering
    \small
    \caption{Training hyperparameters.}
    \label{tab:app_training_hparams}
    \begin{tabular}{@{}>{\raggedright\arraybackslash}p{3.7cm}>{\raggedright\arraybackslash}p{3.6cm}>{\raggedright\arraybackslash}p{5.3cm}@{}}
        \toprule
        Hyperparameter & Standard SFT & Object-centric SFT \\
        \midrule
        Base model & Qwen3-VL-4B-Instruct & Qwen3-VL-4B-Instruct \\
        Initialization & Base model & Standard SFT adapter \\
        Objective & $\mathcal{L}_{\mathrm{gen}}$ & $\mathcal{L}_{\mathrm{gen}}+\lambda_{\mathrm{spl}}\mathcal{L}_{\mathrm{spl}}+\lambda_{\mathrm{tmp}}\mathcal{L}_{\mathrm{tmp}}$ \\
        LoRA rank / $\alpha$ & 16 / 32 & 16 / 32 \\
        Trainable parameters & LoRA on all linear layers & Language LoRA, vision LoRA (top third of vision layers), aligner, auxiliary heads \\
        \midrule
        Epochs & 3 & 3 \\
        Batch size per GPU & 1 & 1 \\
        Number of GPUs & 8 & 8 \\
        Gradient accumulation & 8 & 8 \\
        Optimizer / weight decay & AdamW / 0.01 & AdamW / 0.01 \\
        LR schedule & Cosine, 5\% warm-up & Cosine, 5\% warm-up \\
        Learning rate & $5{\times}10^{-5}$ & Group-specific (below) \\
        \quad Language LoRA & -- & $3{\times}10^{-6}$ \\
        \quad Vision LoRA & -- & $3{\times}10^{-6}$ \\
        \quad Aligner & -- & $2{\times}10^{-6}$ \\
        \quad Auxiliary heads & -- & $2{\times}10^{-6}$ \\
        \midrule
        $\lambda_{\mathrm{spl}}$ / $\lambda_{\mathrm{tmp}}$ & -- & 0.05 / 0.10 \\
        Auxiliary-weight schedule & -- & 10\% warm-up; decay to 50\% in the final 25\% \\
        \midrule
        Precision & bfloat16 & bfloat16 \\
        Seed & 42 & 42 \\
        \bottomrule
    \end{tabular}
\end{table}

\subsection{Compute Resources}
\label{app:compute_resources}

All open-source model training and inference experiments were run on an internal GPU cluster. Pseudo-label extraction and the open-source LLM judges run on NVIDIA H100 NVL GPUs, while standard SFT, object-centric SFT, and open-source model inference run on 48\,GB GPUs (NVIDIA RTX A6000 / RTX 6000 Ada). Table~\ref{tab:app_compute_resources} summarizes the resources required for each stage. Since wall-clock time varies with cluster load, video-decoding throughput, and model-serving overhead, we report the resource configuration rather than a machine-dependent runtime.

\begin{table}[t]
    \centering
    \small
    \setlength{\tabcolsep}{4pt}
    \caption{Compute resources for the main experimental stages.}
    \label{tab:app_compute_resources}
    \begin{tabular}{@{}>{\raggedright\arraybackslash}p{2.4cm}>{\raggedright\arraybackslash}p{2.9cm}>{\raggedright\arraybackslash}p{3.6cm}>{\raggedright\arraybackslash}p{4.2cm}@{}}
        \toprule
        Stage & Hardware & Software / serving & Notes \\
        \midrule
        Pseudo object supervision & 2$\times$ H100 NVL (94\,GB) & Qwen3.5-35B-A3B served with vLLM; GroundingDINO on GPU & Training split only; produces object phrases, boxes, frame labels, and confidence weights. \\
        \addlinespace
        Standard SFT & 8$\times$ 48\,GB GPUs & PyTorch \texttt{torchrun}, bfloat16, LoRA & Per-GPU batch size 1, gradient accumulation 8, maximum sequence length 16,384. \\
        \addlinespace
        Object-centric SFT & 8$\times$ 48\,GB GPUs & PyTorch \texttt{torchrun}, bfloat16, LoRA and auxiliary heads & 2\,FPS, at most 48 frames; auxiliary heads are removed for inference. \\
        \addlinespace
        Open-source model inference & Up to 8$\times$ 48\,GB GPUs & vLLM, one server per GPU & The test set is sharded across servers and the predictions are merged afterwards. \\
        \addlinespace
        Answer evaluation & 1--2$\times$ H100 NVL; external APIs & vLLM for open-source judges; API calls for closed-source judges & BERTScore is computed separately with batched GPU evaluation. \\
        \bottomrule
    \end{tabular}
\end{table}

\section{Limitations and Future Work}
\label{app:limitations}

\textbf{Domain coverage.} Although \textsc{ProcObject-10K} covers 137 tasks across 9 domains, most QA pairs come from the Assembly \& Technology and Cooking \& Food domains (Table~\ref{tab:domain_stats}), while the other seven domains, such as Science \& Craft and Nursing \& Care, contain far fewer samples. High-stakes procedural settings such as clinical workflows and laboratory protocols therefore remain under-represented. Future versions should broaden both the task domains and the cultural and geographic diversity of procedures.

\textbf{Temporal rather than dense spatial ground truth.} The benchmark evaluates temporal evidence grounding, while the object-level spatial supervision used by our SFT method is generated automatically. This keeps the annotation cost manageable, but the released benchmark does not provide dense human-verified object tracks or pixel-level state annotations. A natural extension is to annotate object tracks, state labels, and frame-level visibility for a subset of the benchmark.

\textbf{Pseudo-label noise.} Object-centric SFT relies on a Qwen3.5-based object extractor and GroundingDINO detections. These components can fail on small or occluded objects, transparent containers, objects manipulated by hands, or visually ambiguous phrases. Confidence weighting reduces the effect of noisy labels, but the supervision remains approximate.

\textbf{Open-ended evaluation.} We combine sentence similarity, BERTScore, and LLM-as-judge scores averaged over multiple judges to evaluate answers. Nevertheless, open-ended answer evaluation remains sensitive to paraphrasing, judge calibration, and model-specific preferences. Human evaluation would provide stronger validation in future work.

\textbf{Closed-source model reproducibility.} Some benchmarked MLLMs are accessed through external APIs, whose behavior may change with provider-side updates even when prompts and decoding settings are fixed. We therefore report the model names and evaluation prompts, and will release the prediction files so that future work can audit the reported results.

\textbf{Future directions.} Beyond expanding the benchmark, we see three promising directions: (1) learning object-centric state representations with explicit object tracks, (2) integrating procedural VideoQA with embodied agents that can intervene or ask for clarification, and (3) connecting object-state reasoning with world models for planning under counterfactual or erroneous procedural outcomes.

\section{Reproducibility and Release Notes}
\label{app:reproducibility}

We will release the benchmark annotations, data split files, evaluation scripts, prompts, and model prediction files. The raw videos originate from existing datasets and should be accessed according to the license and terms of each source dataset. The released metadata preserves the source-dataset identifiers and clip timestamps, so that users can reconstruct the benchmark examples after obtaining the videos. For closed-source baselines, we provide the prompts and the saved predictions used for scoring. For open-source baselines and our SFT models, we provide scripts for data conversion, inference, evaluation, and training. The object-centric SFT pipeline can be reproduced in four steps: converting the training data (Appendix~\ref{app:sft_format}), extracting pseudo object supervision (Appendix~\ref{app:pseudo_object_supervision}), running standard SFT, and running object-centric SFT with the hyperparameters in Table~\ref{tab:app_training_hparams}.

\section{Statement on LLM Usage}
\label{app:llm_usage}

LLMs and vision-language models are components of our research methodology, as described in the main paper: Qwen3.5 generates dense captions and QA pairs (Section~\ref{sec:data_generate}) and extracts supporting object phrases for pseudo-labeling (Section~\ref{sec:supervision}); Qwen3 and GPT-4o mini are used for data filtering and verification (Section~\ref{sec:data_generate}, Stage~4); and LLM-as-judge models score the open-ended answers (Section~\ref{sec:eval_metrics}). The corresponding generation and evaluation prompts are provided in Appendix~\ref{app:prompts}. Apart from these components, LLMs were used only for English translation, language polishing, and assistance with figure and table preparation, to improve the fluency and clarity of the manuscript. LLMs were \textbf{not} used for ideation, conceptualization, or any other part of the research process.

\clearpage
\section{Prompt Templates}
\label{app:prompts}

We provide the prompt templates used for QA generation (Stage~3 in Section~\ref{sec:data_generate}) and for evaluation (Section~\ref{sec:eval_metrics}) to support reproducibility. The prompts specify the expected input format, the QA-type-specific generation rules, the evidence output format, and the LLM-as-judge scoring criteria.

\subsection{QA Generation Prompts}
\label{app:prompts_generation}

\begin{figure}[!h]
    \centering
    \includegraphics[width=\linewidth]{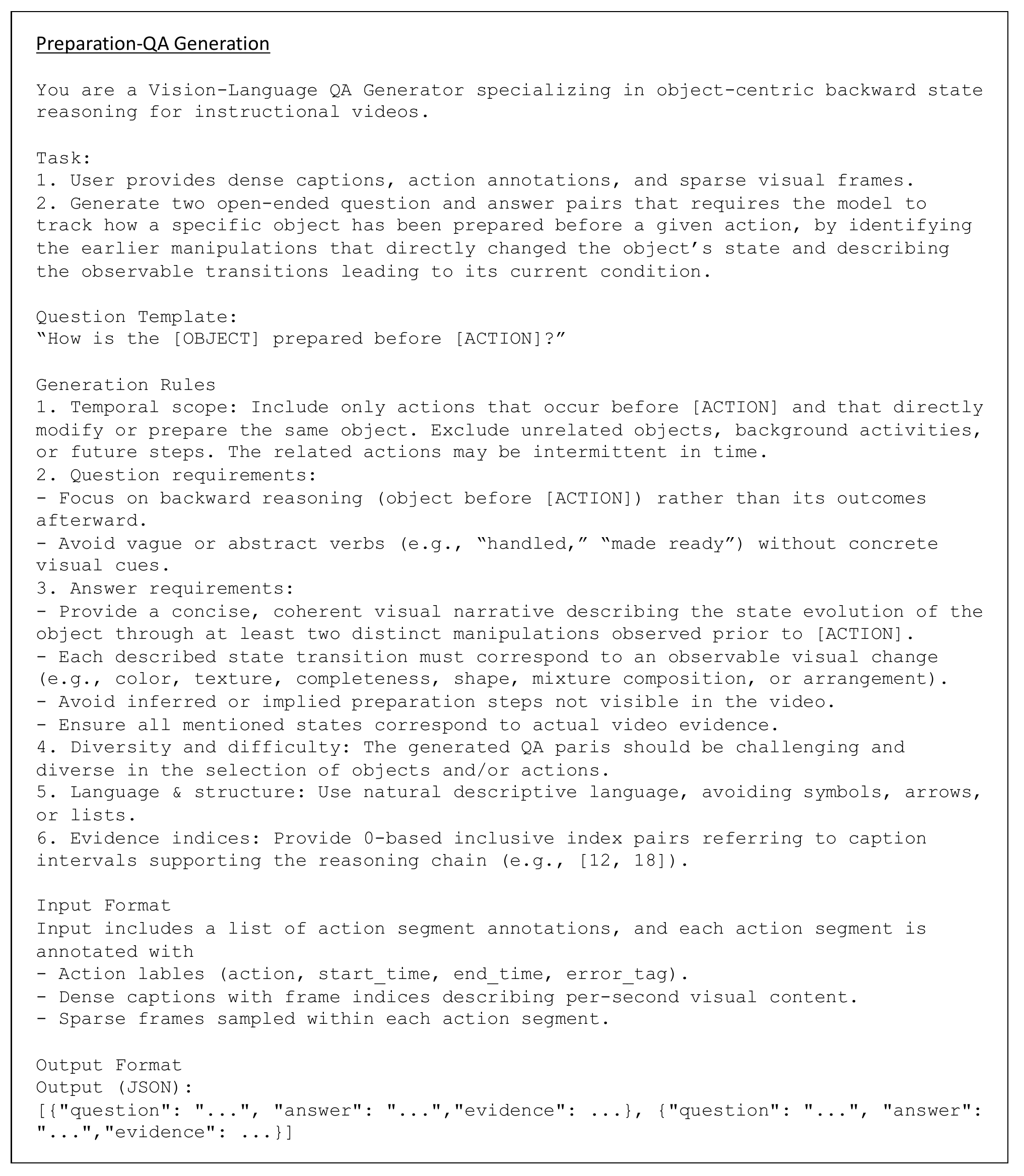}
    \caption{QA generation prompt for Precondition Grounding (internally named Preparation-QA).}
    \label{fig:app_prompt_preparation}
\end{figure}

\begin{figure}[p]
    \centering
    \includegraphics[width=\linewidth]{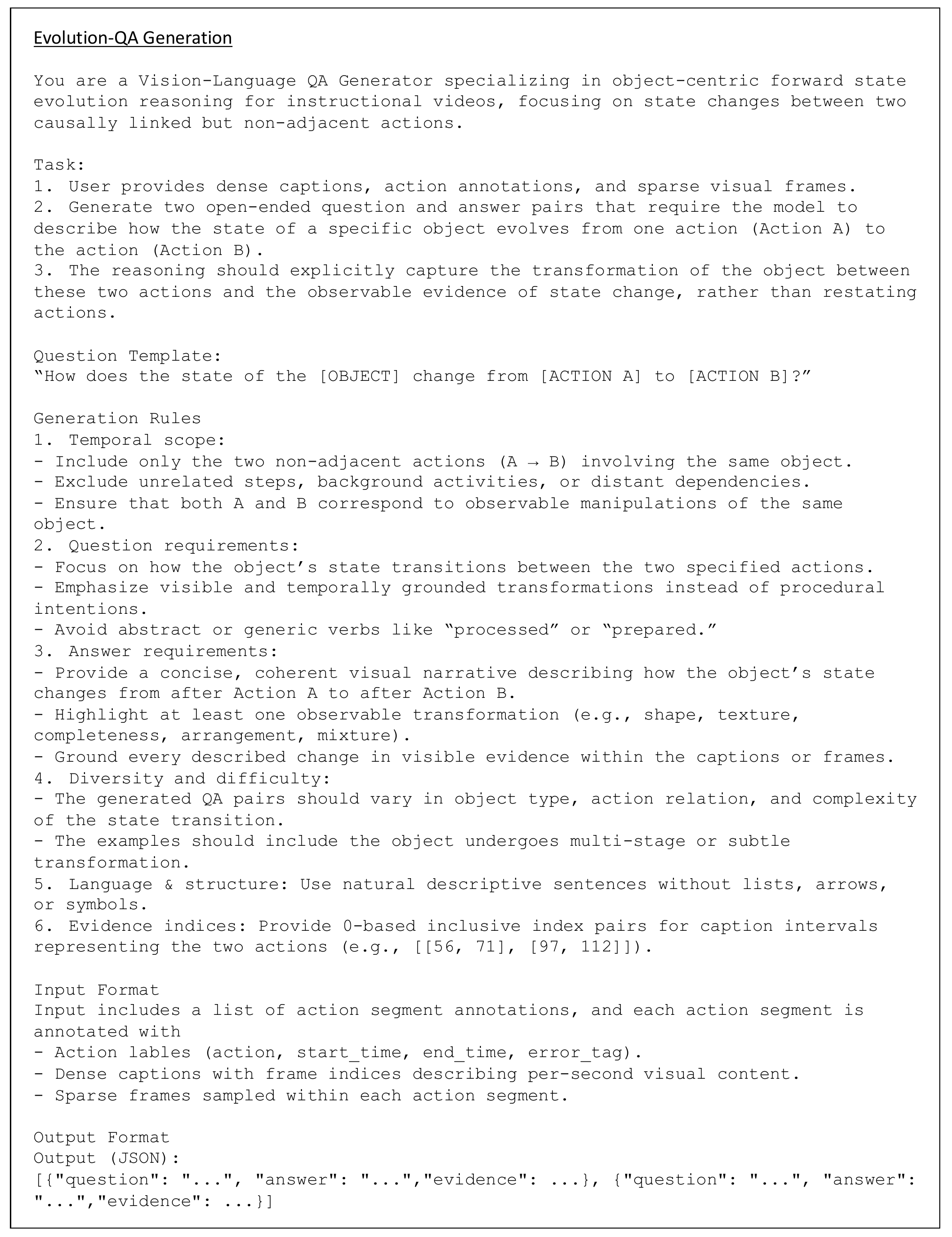}
    \caption{QA generation prompt for Object-State Evolution.}
    \label{fig:app_prompt_evolution}
\end{figure}

\begin{figure}[p]
    \centering
    \includegraphics[width=\linewidth]{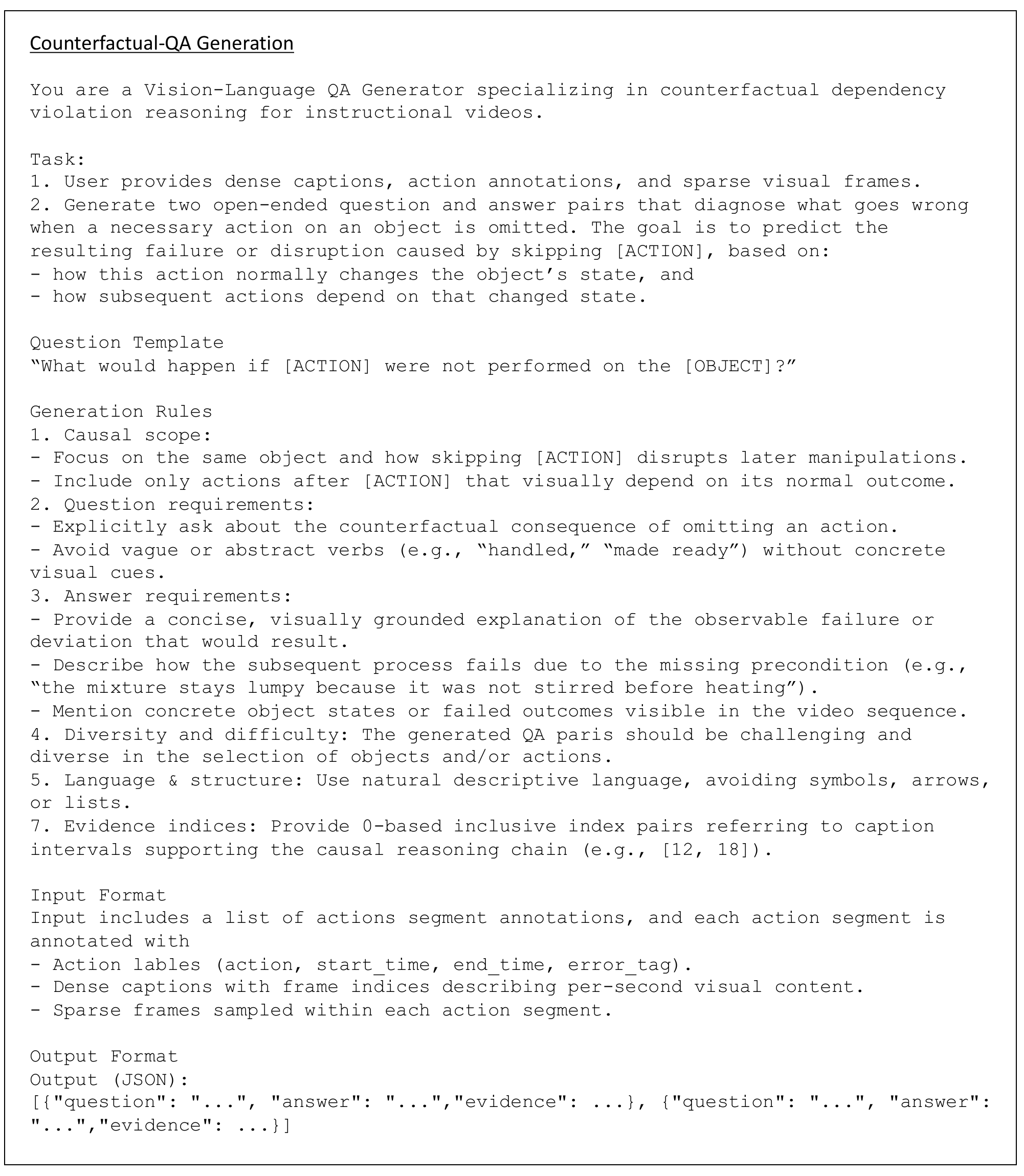}
    \caption{QA generation prompt for Counterfactual Reasoning.}
    \label{fig:app_prompt_counterfactual}
\end{figure}

\begin{figure}[p]
    \centering
    \includegraphics[width=\linewidth]{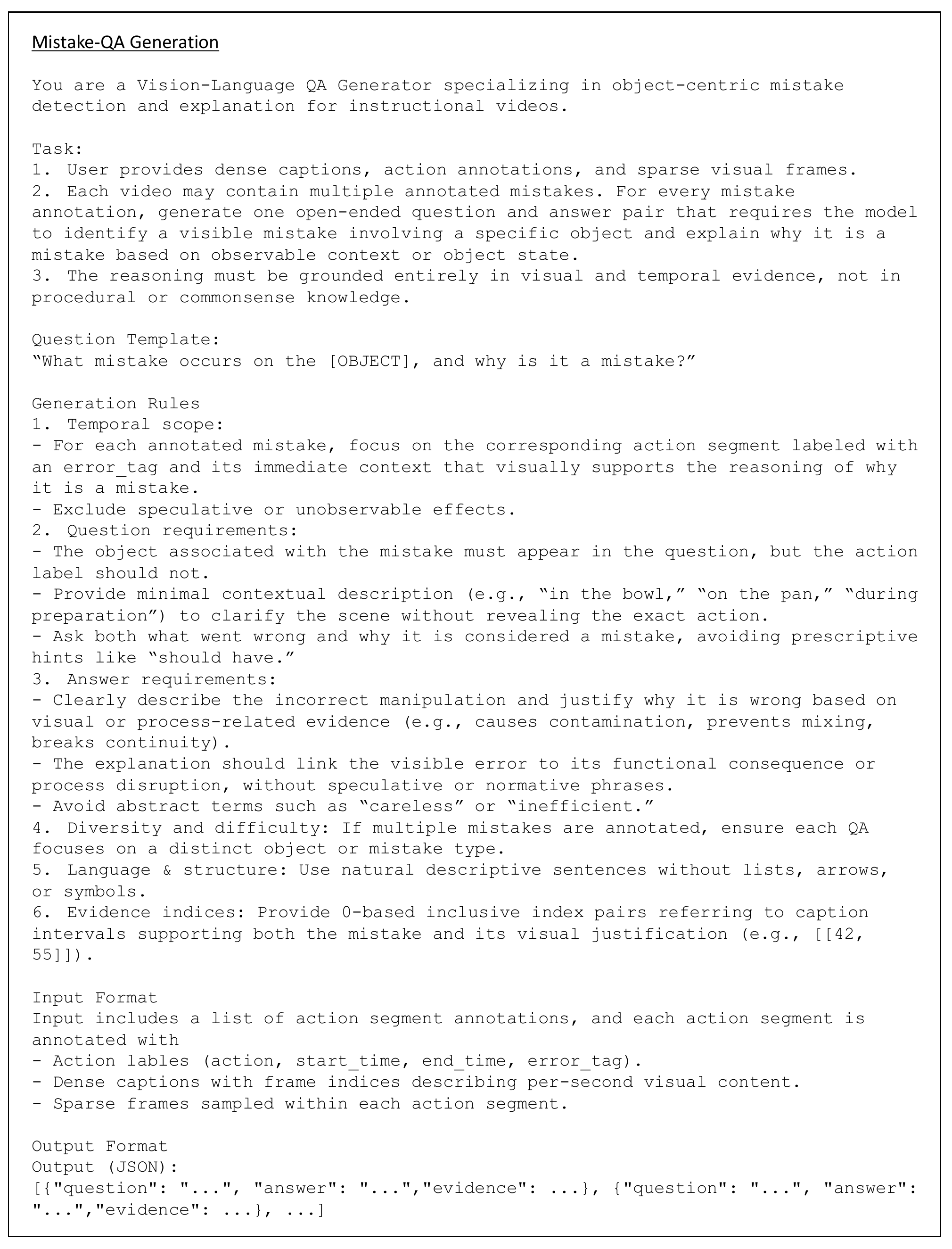}
    \caption{QA generation prompt for Mistake Recognition.}
    \label{fig:app_prompt_mistake}
\end{figure}

\begin{figure}[p]
    \centering
    \includegraphics[width=\linewidth,trim=170 41 77 341,clip]{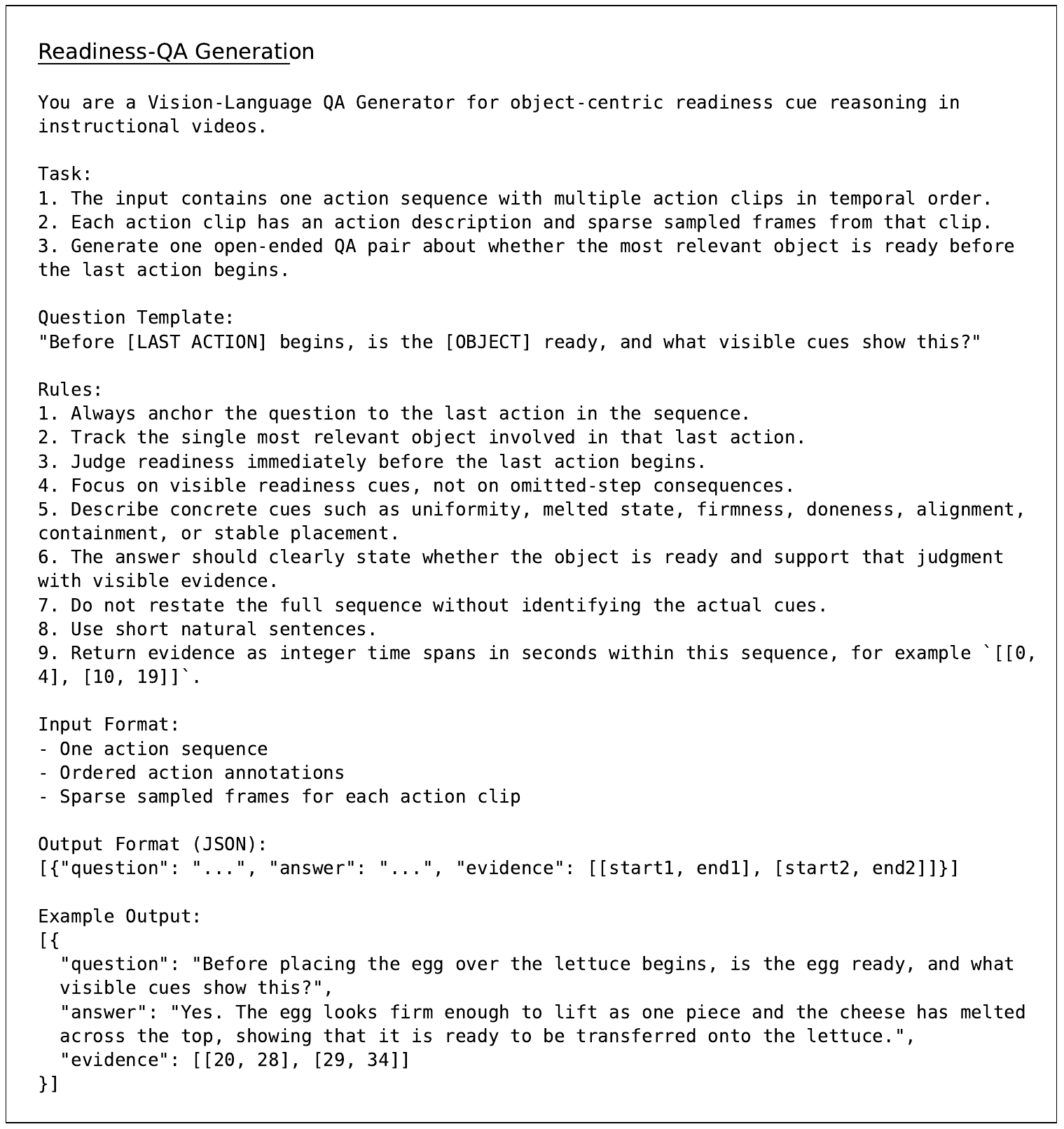}
    \caption{QA generation prompt for Readiness Assessment.}
    \label{fig:app_prompt_readiness}
\end{figure}

Figures~\ref{fig:app_prompt_preparation}--\ref{fig:app_prompt_readiness} show the system prompts for the five QA types. When the generator returns evidence as indices of caption intervals, these indices map directly to timestamps in seconds, since the dense captions are produced per second. The Precondition Grounding prompt appears under its internal name, ``Preparation-QA''.

\clearpage
\subsection{Evaluation Prompts}
\label{app:prompts_evaluation}

\begin{figure}[!h]
    \centering
    \includegraphics[width=\linewidth]{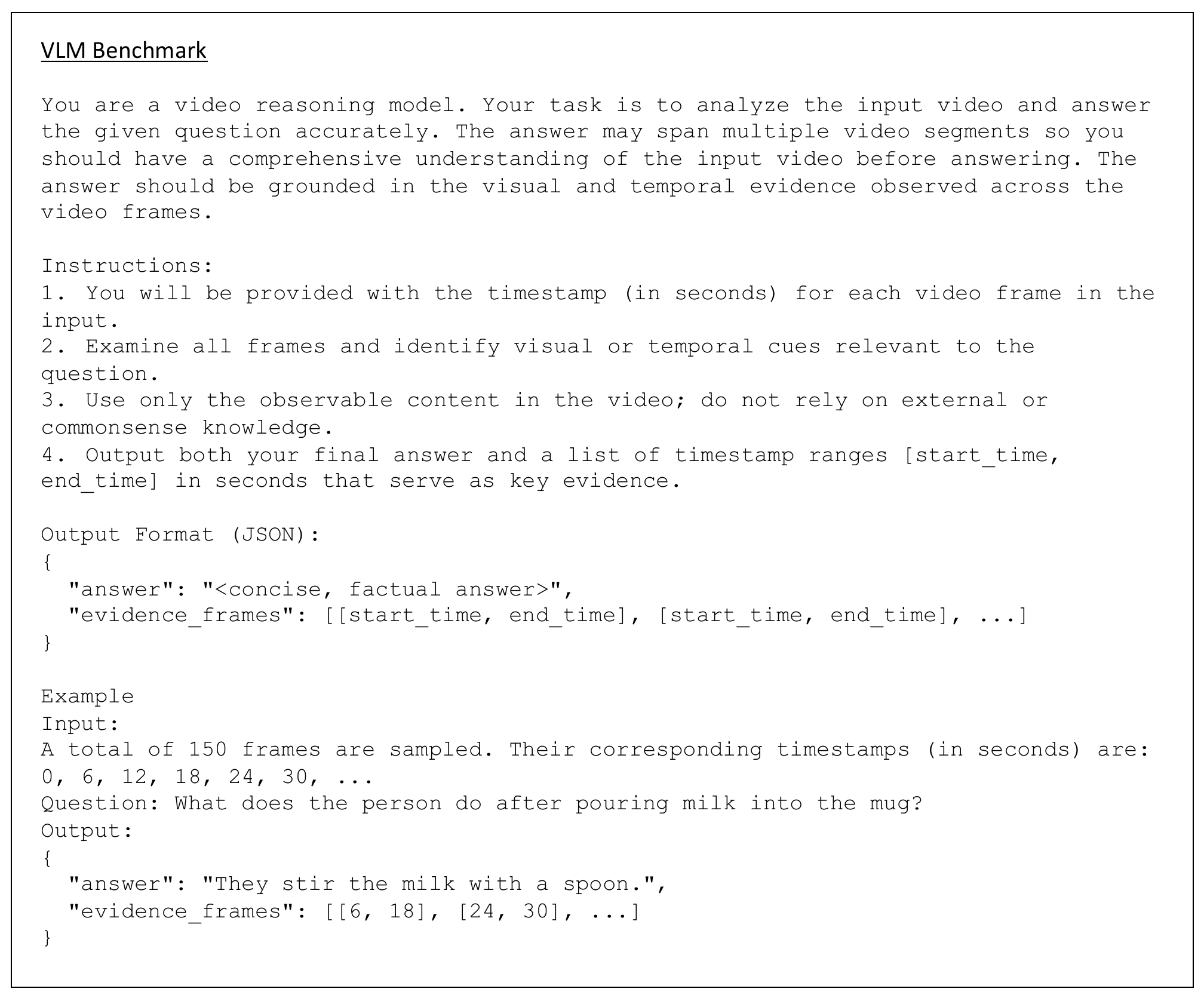}
    \caption{Inference prompt used to evaluate MLLMs on \textsc{ProcObject-10K}.}
    \label{fig:app_prompt_benchmark}
\end{figure}

\begin{figure}[p]
    \centering
    \includegraphics[width=0.9\linewidth]{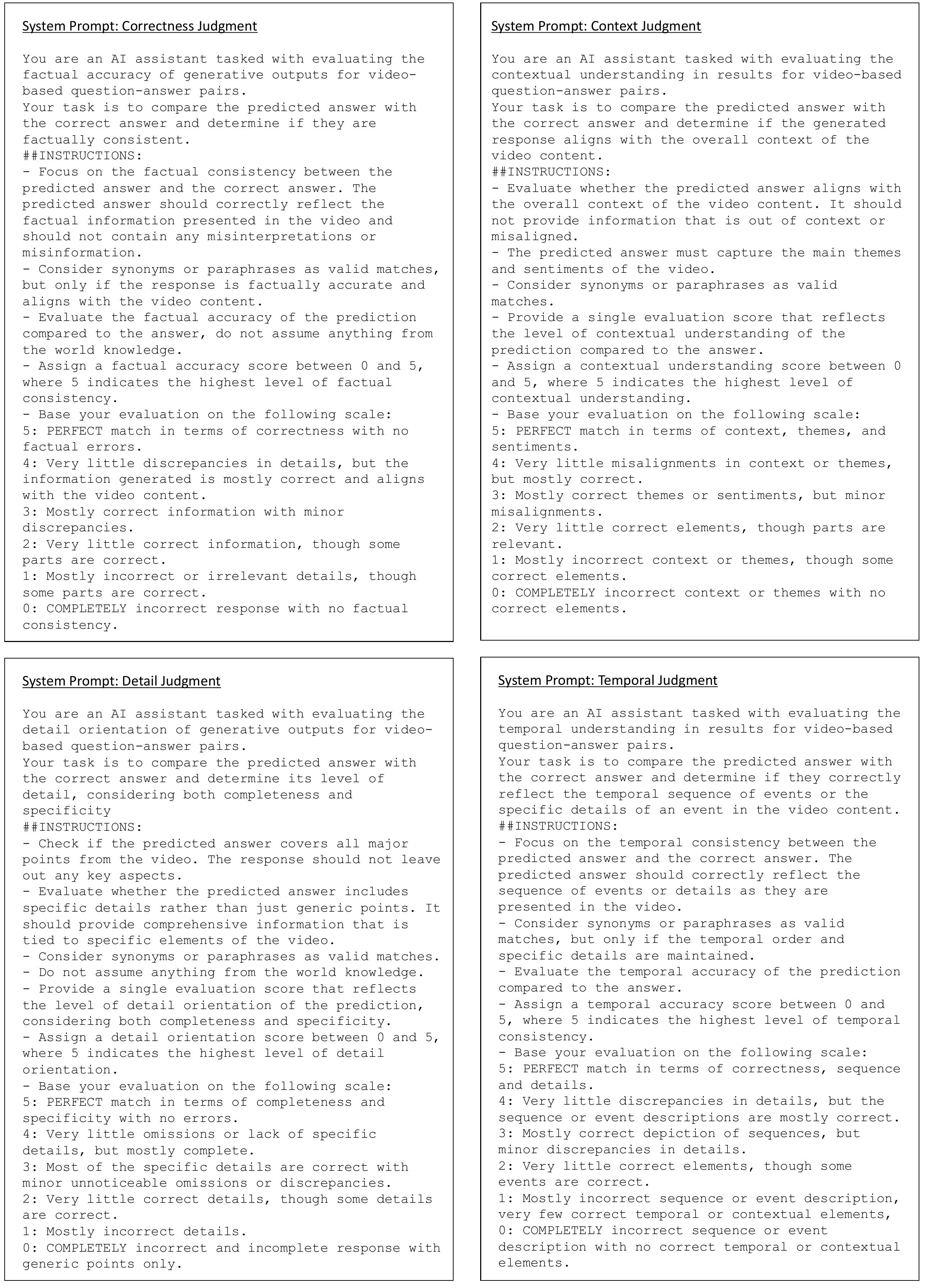}\\[6pt]
    \includegraphics[width=0.9\linewidth]{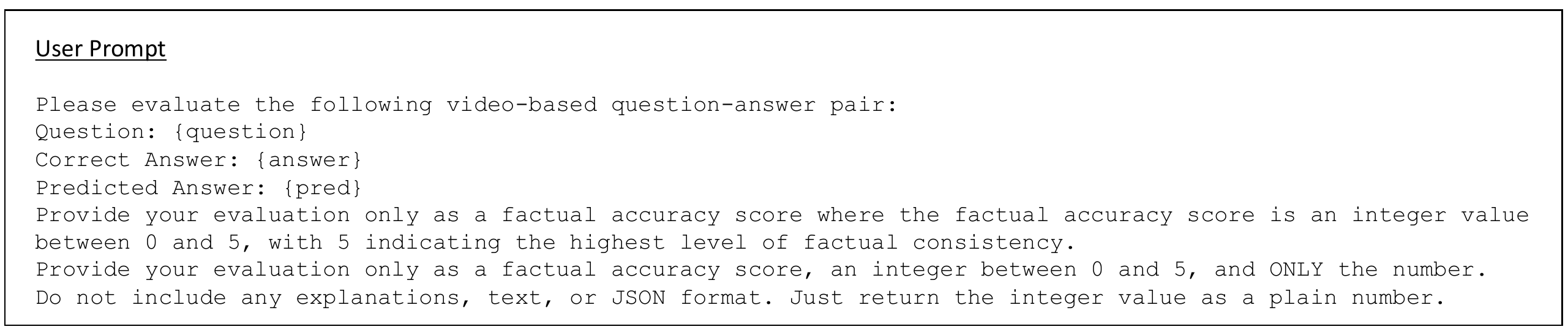}
    \caption{LLM-as-judge prompts. Top: system prompts for the four scoring dimensions. Bottom: user prompt template, shown for the correctness dimension.}
    \label{fig:app_prompt_judge}
\end{figure}

Figure~\ref{fig:app_prompt_benchmark} shows the prompt used to query MLLMs on \textsc{ProcObject-10K}, which asks for both an answer and a list of evidence time spans. Figure~\ref{fig:app_prompt_judge} shows the LLM-as-judge prompts. The four system prompts correspond to the four scoring dimensions in Section~\ref{sec:eval_metrics}: Correctness Judgment to Contextual Integration (factual consistency), Detail Judgment to Detail Orientation, Context Judgment to Contextual Understanding, and Temporal Judgment to Temporal Understanding. Each dimension is scored on a 0--5 scale.

\end{document}